# PeruMedQA: Benchmarking Large Language Models (LLMs) on Peruvian Medical Exams - Dataset Construction and Evaluation

Rodrigo M. Carrillo-Larco[1,2]

Jesus Lovón Melgarejo[3]

Manuel Castillo-Cara[4,5]

Gusseppe Bravo-Rocca[6]

1. Hubert Department of Global Health, Rollins School of Public Health, Emory University, Atlanta, GA, USA.
2. Emory Global Diabetes Research Center of Woodruff Health Sciences Center, Emory University, Atlanta, USA.
3. Institut de Recherche en Informatique de Toulouse, Toulouse, France.
4. Universidad Nacional de Educación a Distancia, Madrid, Spain.
5. Instituto de Investigación Científica, Universidad de Lima, Lima, Peru.
6. Barcelona Supercomputing Center, Barcelona, Spain.

**Corresponding author**

Rodrigo M Carrillo-Larco, MD, PhD

Rollins School of Public Health, Emory University, Atlanta, GA, USA.

rmcarri@emory.edu

## ABSTRACT

**BACKGROUND**: Medical large language models (LLMs) have demonstrated remarkable performance in answering medical examinations. However, the extent to which this high performance is transferable to medical questions in Spanish and from a Latin American country remains unexplored. This knowledge is crucial as LLM-based medical applications gain traction in Latin America. **AIMS**: To build a dataset of questions from medical examinations taken by Peruvian physicians pursuing specialty training; to fine-tune a LLM on this dataset; to evaluate and compare the performance in terms of accuracy between vanilla LLMs and the fine-tuned LLM. **METHODS**: We curated PeruMedQA, a multiple-choice question-answering (MCQA) dataset containing 8,380 questions spanning 12 specialties (2018-2025). We selected ten medical LLMs, including medgemma-4b-it and medgemma-27b-text-it, and developed zero-shot task-specific prompts to answer the questions. We employed parameter-efficient fine tuning (PEFT) and low-rank adaptation (LoRA) to fine-tune medgemma-4b-it utilizing all questions except those from 2025 (test set). **RESULTS**: Medgemma-27b showed the highest accuracy across all specialties, achieving the highest score of 89.29% in Psychiatry; yet, in two specialties, OctoMed-7B exhibited slight superiority: Neurosurgery with 77.25% and 77.38%, respectively; and Radiology with 76.13% and 77.30%, respectively. Across specialties, most LLMs with <10 billion parameters exhibited <50% of correct answers. The fine-tuned version of medgemma-4b-it emerged victorious against all LLMs with <10 billion parameters and rivaled a LLM with 70 billion parameters across various examinations. **CONCLUSIONS**: For medical AI applications and research that require knowledge bases from Spanish-speaking countries and those exhibiting similar epidemiological profiles to Peru's, interested parties should utilize medgemma-27b-text-it.

## INTRODUCTION

Large language models (LLMs) have transformed generative artificial intelligence (AI) across various creative and factual tasks in diverse domains, including medicine.[1-5] Notably, when trained on medical data, LLMs have shown high performance in medical exams, producing outputs comparable or superior to those of health professionals.[6-9]

Despite these advancements, the capabilities of LLMs in answering medical questions in Spanish remain largely unexplored.[10] Furthermore, the ability of medical LLMs to accurately answer medical questions from countries in South America with unique epidemiological profiles, which combine noncommunicable chronic diseases with infectious and tropical diseases, is also unknown. Notably, these countries often exhibit distinct epidemiological patterns[11,12] compared to high-income countries from which most training datasets have been sourced.[6,13] Additionally, previous evidence suggests that LLMs that perform well in medical questions from the United States[6] experience performance degradation when tasked with questions from Brazil[14] and the African continent.[15]

Consequently, it is crucial to identify the most effective LLMs to answer medical questions in Spanish from South American countries, as more LLM-based medical applications emerge and become prevalent in Latin America. By doing so, these applications can utilize the LLMs that provide the most accurate responses. To deliver this evidence, we initially constructed a dataset of medical questions from the examinations that medical professionals in Peru undertake to pursue specialty and subspecialty training.

Subsequently, we assigned ten medical LLMs to respond to these questions. Finally, we fine-tuned one LLM and evaluated whether this procedure enhanced its accuracy. Throughout our examination, we utilized LLMs ranging from 4 billion (4B) to 70 billion (70B) parameters; intuitively, the greater the number of parameters, the more capable the LLM becomes. Thus, we anticipate that LLMs with a higher number of parameters will exhibit superior performance.

## METHODS

### Dataset

#### *Data Source*

The *Consejo Nacional de Residentado Médico* (CONAREME; National Council of Medical Residencies) is the official institution in Peru responsible for preparing and administering the medical examinations required for medical professionals seeking further training.[16] Following each selection process, CONAREME publishes on its website the examinations along with the correct answers.[16] These examinations are multiple-choice assessments, and all questions are formulated in Spanish.

CONAREME prepares the exams for both general specialty training (e.g., a graduated medical doctor aspiring to become a gastroenterologist) and subspecialty training (e.g., a medical doctor with specialty training in pediatrics seeking further specialization in pediatric gastroenterology).[16] This latter form of training, commonly referred to as post-residency, is equivalent to fellowship training in the United States.

### *Data Processing*

We downloaded PDF files containing the examinations and the correct answers from the years 2018, 2019, 2020, 2022, 2023, 2024, and 2025 (Supplementary Table 1). The exams for the year 2021 were not available on the CONAREME website.[16] To facilitate the extraction of exam questions, possible answers, and correct answers, we developed Python programs.[17] This code read each PDF file and extracted the relevant information, which was subsequently organized and saved as a dataset in CSV files. This process was conducted individually for each examination across the specified years and specialty as well as subspecialties.

### *Manual Verification*

To ensure the accuracy of the extracted information, one human (RMC-L) verified that the correct answers had been correctly identified from the original PDFs. If the correct answer extracted by the Python program was incorrect, we manually updated the corresponding CSV file. This manual correction was necessary for 16 questions out of the 8,380 total questions (Supplementary Table 1). Thus, our programmatic approach to extracting correct answers was mostly effective,[17] with 0.19% (16/8,380) of all possible cases being incorrect. We also verified that when there were numbers in the multiple-choice answers, these were correctly formatted and were not mistakenly interpreted as dates by the CSV files. We did not verify the correctness of the questions or the multiple-choice answers.

### *Postprocessing*

In certain years (2023, 2024, and 2025), the examinations presented four multiple-choice answers (A, B, C, and D). Conversely, in other years (2018, 2019, 2020, and 2022), the examinations offered five possible answers (A, B, C, D, and E).[16] To ensure uniformity in the number of possible answers an additional option labeled "NA" (none of the above) was introduced to the questions which originally had only four options. This was key to ensure that all tests had an equal number of options, thereby maintaining consistent complexity. A test with four options would be easier than one with five options. In the former scenario, any answer would have a 25% probability of being correct, whereas in the latter, it would have a 20% probability. Finally, to safeguard the special characters present in the Spanish language, the dataset was saved and subsequently utilized as a pickle file. We named this dataset PeruMedQA.

### *Nomenclature*

The examinations designated as Test A and Test B are intended for medical professional pursuing specialty training, such as general physicians who opt for cardiology training. Examinations labeled with specific medical fields, such as pediatrics, are designed for subspecialty or fellowship training. For instance, a general physician who has already completed specialty training in pediatrics may seek further subspecialty training in a specific field like pediatric cardiology.

## LLMs answer medical questions

### *Approach*

In accordance with the standard methodology employed in comparable analytical frameworks,[6-9,14,15] we evaluated the efficacy of several LLMs in answering medical questions in Spanish derived from the CONAREME examinations. We developed a zero-shot task-specific prompt (Supplementary Table 2), wherein the system and user messages were composed in Spanish. The user message encompassed the question, accompanied by multiple-choice options. The LLMs were tasked with responding to the questions by selecting a single option from the available choices.

### *Selected LLMs*

We exclusively selected LLMs specialized in the medical domain and opted for small LLMs, all with a parameter count less than 10 billion (B). This decision was driven by the need to minimize computational resources, as future research and practical applications in Peru and other settings with limited computational resources would likely benefit from smaller LLMs. Despite this rationale, we also utilized two larger LLMs: one with 27 billion (27B) parameters and another with 70 billion (70B) parameters. We hypothesized that they would yield the highest accuracy, and we therefore used the 27B and 70B to portray an ideal scenario.

Specifically, we employed ten LLMs (Supplementary Table 3): MediPhi-Instruct-3.8B, medgemma-4b-it, medgemma-1.5-4b-it, BioMistral-7B-DARE, meditron-7b, OctoMed-7B, Llama3-OpenBioLLM-8B, JSL-MedLlama-3-8B-v2.0, medgemma-27b-text-it, and Llama3-OpenBioLLM-70B. These LLMs are accessible on Hugging Face and can be seamlessly integrated into Python programming through the Transformers library.

Notably, the "B" at the end of the names of the LLMs refer to billions, which indicates the number of parameters each LLM internally possesses. For instance, medgemma-4b-it has four billion parameters. Furthermore, the greater the number of parameters, the more capable the LLM is expected to be. Consequently, medgemma-27b-text-it with 27 billion parameters is expected to yield superior performance than medgemma-4b-it (four billion parameters), while Llama3-OpenBioLLM-70B should outperform other LLMs due to its 70 billion parameters.

***Evaluation***

The prepared datasets, as outlined earlier, served as the gold standard for evaluating the performance of the LLMs. That is, the responses provided by the LLMs were evaluated against the answers in the tests. The LLMs were prompted to respond to medical questions by selecting one of the multiple-choice answers. In other words, the LLMs were prompted to respond to medical questions by selecting one of the multiple-choice answers, following a similar approach to the Massive Multitask Language Understanding (MMLU) benchmark, which evaluates language models across diverse academic subjects using multiple-choice questions.[18]

We compared the answers provided by the LLM with the correct answers from the CONAREME examinations. This paper presents descriptive statistics overall and stratified by LLM and test as well as year. The year stratification becomes pertinent in verifying whether LLMs consistently deliver comparable accuracies across different years. Furthermore, despite these examinations being standardized, the questions may undergo modifications, and the topics may also change. For instance, COVID-19-related questions were absent in 2018. Also, certain topics may acquire greater prominence and correspondingly, questions may be formulated based on the underlying epidemiological profile. For example, questions about Malaria may become more prevalent during years characterized by severe outbreaks. This is a descriptive analysis.

In certain cases, the LLMs hallucinated and provided invalid answers ignoring the prompt instructions (Supplementary Table 4). The term “hallucination” pertains to instances where the LLM exhibits unexpected behavior. For instance, they may have provided an alternative answer when there were only five options (e.g., K), or they may have explained their underlying reasoning without providing a letter answer as requested. Consequently, some exams were not fully answered. This is equivalent to missing data or missing answers. Therefore, we computed the percentage of correctly answered questions as the number of questions correctly answered by the LLM divided by the number of questions with valid answers.

In addition to computing the percentage of correctly answered questions, we investigated the percentage of questions correctly answered by each LLM. The denominator was the total number of available questions (8,380), and the numerator was the number of questions that each model answered correctly but all other LLMs answered incorrectly. For example, the percentage of questions that *only* medgemma-4b-it answered correctly.

## Fine-tuning

### *Overview*

We selected the most recent and compact medical LLM available at the time of writing for parameter-efficient fine-tuning (PEFT) employing Low-Rank Adaptation (LoRA). Specifically, we fine-tuned the model medgemma-4b-it.[13] LoRA's primary function is to represent the weights update of a LLM without updating the model. LoRA trains smaller matrices to represent the necessary adjustments, effectively capturing the updates in a low-rank form. In simpler terms, LoRA extends and combines the internal layers of a LLM for efficient fine-tuning on a new task. This approach reduces cost in terms of resources and time compared to classical fine-tuning. In this work, the new task or domain was medical questions from the Peruvian examinations formulated in Spanish.[16]

### *Experimental setup*

We utilized the same dataset we prepared and detailed earlier. We reserved the questions from the 2025 examinations as the test set (1,400 questions). The remaining dataset (excluding the 2025 year) was divided into training (90% or 6,282 questions) and validation (10% or 698 questions). We conducted a training process spanning 10 epochs, commencing with a learning rate of $5^{-5}$, the R and alpha parameters for LoRA were 16 and 16, respectively, the LoRA's dropout was 0.05, and the targeted modules were "all-linear". We largely adhered to the instructions provided in the medgemma cookbook for the fine-tuning process.[19] We developed a custom metric for monitoring the training process. This metric was the accuracy, defined as the number of correct answers divided by the total number of valid answers.

### Ethics

We used publicly available exam questions available online.[16] We lacked privileged access to any data and did not engage with human subjects. This work was deemed to pose minimal risk.

### Role of the funding source

There was no specific funding for this work.

## RESULTS

### LLMs answer medical questions – Hallucinations

In this section, we report the findings regarding instances where LLMs failed to provide answers as prompted, a phenomenon known as hallucinations. Specifically, LLMs were instructed to yield a single answer, either A, B, C, D, or E. However, in some cases, LLMs showed hallucinations by failing to provide any answer or by delivering any other letter (e.g., K), regardless of whether the answer was correct or not.

We observed variability in the number of invalid answers provided by the LLMs; in other words, number of answers in which the LLMs hallucinated (Supplementary Table 4). The top three LLMs with the fewest number of hallucinations were Llama3-OpenBioLLM-70B (0.00%), medgemma-27b-text-it (0.02%), and MediPhi-Instruct-3.8B (0.04%). The LLM medgemma-4b-it also exhibited a low number of invalid answers (0.14%), which further improved to 0.00% after fine-tuning. Conversely, LLMs with the largest number of hallucinations were meditron-7b (66.37%), JSL-MedLlama-3-8B-v2.0 (9.00%) and Llama3-OpenBioLLM-8B (4.96%). These proportions do not provide any insights regarding the accuracy of the LLMs (i.e., whether the answers were correct or not), but rather their capacity to adhere to instructions and provide responses as prompted.

**LLMs answer medical questions – Accuracy**

Unlike the previous subsection where we reported how well the LLMs yielded an answer as instructed, regardless of whether the answer was correct or not, in this section we report the results of how well the answers from the LLMs matched the correct answer in the tests (i.e., accuracy or correctness of the LLMs to sit medical tests).

Over the course of several years, three LLMs achieved >85% accuracy in their responses (Figure 1 and Supplementary Table 5). These LLMs were medgemma-27b-text-it (Psychiatry: 89.29%, Test B: 87.44%); Llama3-OpenBioLLM-70B (Psychiatry: 85.57%); and OctoMed-7B (Psychiatry: 85.29%). Conversely, meditron-7b consistently showed the lowest proportion of correct answers (e.g., Pathology and Anatomical Pathology, 19.62%).

Upon further examination of year-specific tests, medgemma-27b-text-it and Llama3-OpenBioLLM-70B achieved >90% accuracy (Figure 2 and Supplementary Table 5). For instance, medgemma-27b-text-it achieved 94.00% in Psychiatry 2025 and 92% in Pathology and Anatomical Pathology 2024; similarly, Llama3-OpenBioLLM-70B achieved 92.00% in Psychiatry 2024 and 91.00% in Psychiatry 2020. The lowest accuracy was delivered by BioMistral-7B-DARE in Ophthalmology 2018 (11.34%), followed by meditron-7b in Anesthesiology 2025 (12.00%).

We analyzed the proportion of questions correctly answered by a single LLM (Supplementary Table 6). OctoMed-7B successfully answered 50 out of 8,380 (0.60%) questions that other LLMs failed to answer correctly. medgemma-27b-text-it ranked second with a success rate of 0.55% (i.e., medgemma-27b-text-it answered correctly 46 questions to which all other LLMs failed). Conversely, meditron-7b only answered nine (0.11%) questions to which all other LLMs failed.

We examined the number of questions that no LLM answered correctly. In total, 190 (2.27%) questions were not answered correctly by any LLM. The frequency of questions that none of the LLMs answered correctly has increased in recent years (Supplementary Table 7). For instance, none of the LLMs accurately answered 21.05% of questions in 2025, compared to 14.74% in 2018. When the number of questions that none of the LLMs answered correctly was categorized by medical field (Supplementary Table 8), surgical fields exhibited the highest proportions: none of the LLMs answered correctly 12.63% of the Ophthalmology questions, whereas this percentage for General Surgery as well as Thoracic and Cardiovascular Surgery was 12.63% and 11.05%, respectively.

**Figure 1. Percent (%) of correct answers by test (specialty or subspeciality) and LLM across years.**

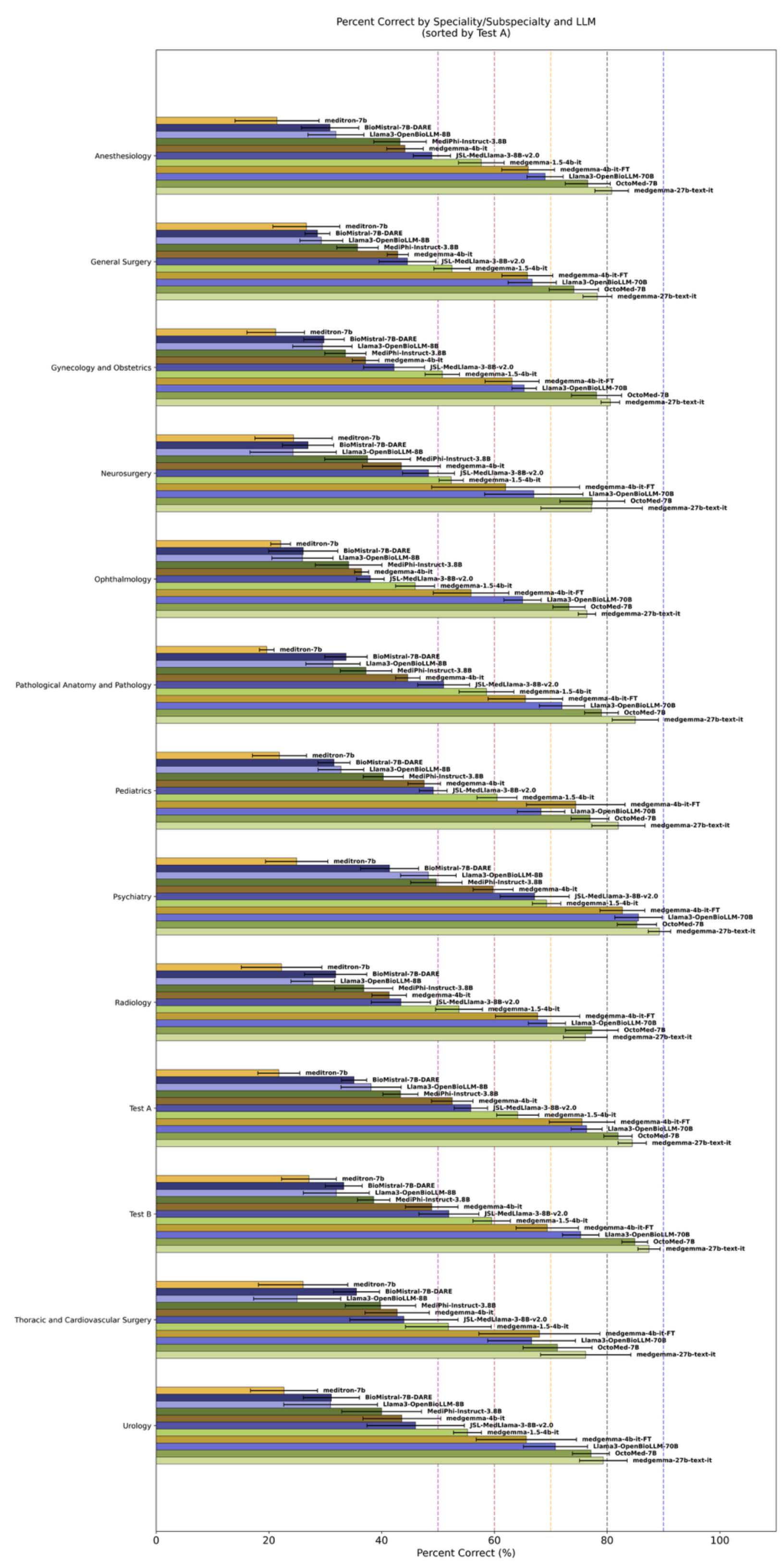

Combined data for all years. The vertical dashed lines serve as visual cues indicating the percentage of correct answers at 50% (purple), 60% (brown), 70% (yellow), 80% (black), and 90% (blue). The color of the bars corresponds to the LLM. The horizontal line at the top of the bar represents the 95% confidence interval. The model medgemma-4b-it-FT refers to the model we fine-tuned, whereas medgemma-4b-it (without the -FT suffix) refers to the vanilla version. The underlying results are shown in Supplementary Table 5.

**Figure 2. Percent (%) of correct answers by test (specialty or subspeciality), LLM and years.**

Percent Correct by Model, Year, and Specialty or Subspecialty

Anesthesiology

General Surgery

Gynecology and Obstetrics

Neurosurgery

Ophthalmology

Pathological Anatomy and Pathology

Pediatrics

Psychiatry

Radiology

Test A

Test B

Thoracic and Cardiovascular Surgery

Urology

BioMistral-7B-DARE
JSL-MedLlama-3-8B-v2.0
Llama3-OpenBioLLM-70B
Llama3-OpenBioLLM-8B
MediPhi-Instruct-3.8B
OctoMed-7B
medgemma-1.5-4b-it
medgemma-27b-text-it
medgemma-4b-it
medgemma-4b-it-FT
meditron-7b

Radar plots are arranged in alphabetical order. Within each radar plot corresponding to a specialty or subspecialty, the examination years are sequentially ordered from the earliest to the most recent available data. Circular lines within the radar plots indicate 20%, 40%, 60%, and 80% of the correct answer percentage. The model medgemma-4b-it-FT refers to the model we fine-tuned, whereas medgemma-4b-it (without the -FT suffix) refers to the vanilla version. The underlying results are shown in Supplementary Table 5.

**Fine-tuned version of medgemma-4b-it**

The fine-tuned version of medgemma-4b-it exhibited superior performance compared to the base medgemma-4b-it model and other LLMs with <10 billion parameters (Figure 1). Also, the fine-tuned version of medgemma-4b-it showed comparable or superior performance to Llama3-OpenBioLLM-70B, yet worse performance than OctoMed-7B and medgemma-27b-text-it in most scenarios (Figure 1 and Figure 2).

For instance, in Neurosurgery 2024, both the fine-tuned version of medgemma-4b-it and Llama3-OpenBioLLM-70B achieved an accuracy of 75.00%, while OctoMed-7B yielded 78.79% and medgemma-27b-text-it scored 83.00% (Supplementary Table 5). In the 2024 Pediatrics test, the fine-tuned version of medgemma-4b-it defeated Llama3-OpenBioLLM-70B by a thirteen-percentage-point margin, achieving a score of 83.00% compared to Llama3-OpenBioLLM-70B's 70.00%. In the 2024 Pediatrics test, the fine-tuned version of medgemma-4b-it also outperformed OctoMed-7B (78.00%) and came close to medgemma-27b-text-it (86.00%). A case in which these four LLMs performed comparably was Radiology 2024: fine-tuned medgemma-4b-it scored 74.00%, Llama3-OpenBioLLM-70B yielded 72.00%, OctoMed-7B achieved 72.00%, and medgemma-27b-text-it got 76.00%. In only three tests did the fine-tuned version of medgemma-4b-it outperform medgemma-27b-text-it: Pediatrics 2022, where medgemma-4b-it fine-tuned outperformed by one percentage point (86.00% vs 85.00%); Test A 2019, where medgemma-4b-it fine-tuned outperformed by one percentage point (83.00% vs 82.00%); and Test A 2023, where medgemma-4b-it fine-tuned outperformed by two percentage points (86.00% vs 84.00%; Supplementary Table 5).

Finally, out of the total number of questions (n=8,380), the fine-tuned version of medgemma-4b-it correctly answered 60 (0.72%) questions for which all the other LLMs provided an incorrect answer (Supplementary Table 6). Notably, this was the highest number of questions uniquely answered by any of the LLMs evaluated in this study.

## DISCUSSION

### Main Findings

Medical LLMs with fewer than 10 billion parameters demonstrated limited success in examinations taken by medical doctors seeking specialty and subspecialty training in Peru. These models predominantly achieved accuracy of <60%, with notable exceptions such as OctoMed-7B which obtained accuracies approaching 90%. The most proficient LLM was medgemma-27b-text-it, which achieved accuracies >90%. A fine-tuned version of medgemma-4b-it substantially enhanced its performance, surpassing the scores of another large model, Llama3-OpenBioLLM-70B; nevertheless, the fine-tuned version of medgemma-4b-it still fell short of medgemma-27b-text-it in all but three tests.

Given the findings, we recommend utilizing MedGemma-27b-text-it for medical applications and research in Peru, as well as in other Spanish-speaking countries where there may be a similar epidemiological profile or disease distribution. A fine-tuned version of a smaller model, as herein demonstrated with medgemma-4b-it, or a similarly capable small LLM such as OctoMed-7B, could also be an effective and efficient alternative. However, it is important to note that utilizing LLMs for research or developing medical AI applications is not a trivial task. We strongly recommend any interested party to comprehensively evaluate the underlying LLMs in their specific task, carefully test their application in real-world scenarios, and weigh the pros and cons, as well as potential benefits against risks, before launching AI-based and LLM-powered medical applications learning on our findings.

**Implications**

We documented that smaller, older models encounter difficulties in adhering to instructions, as evidenced by the substantial number of invalid responses (hallucinations). This observation aligns with the previous findings on the performance of LLMs of varying sizes.[20]

A 27 billion LLM exhibited superior performance than a 70 billion model. Despite the latter's larger size, the former likely outperformed the other due to its recent development and state-of-the-art training.[13] This training utilized extensive data and encompassed various medical disciplines.[13] This finding aligns with prior evaluations of medgemma-27b-text-it, which achieved superior scores compared to Llama3-OpenBioLLM-70B on many medical benchmarks, including MedQA, where medgemma-27b-text-it achieved a 9.5 percentage point advantage.[13] This suggests that larger models do not necessarily yield perfect results, while intermediate models, such as the 27 billion LLM trained with more comprehensive data and recent advancements, may achieve superior performance. This conjecture also applies to our findings where OctoMed-7B, a seven billion model, ranked behind medgemma-27b-text-it in most cases and in two specialties, OctoMed-7B outranked medgemma-27b-text-it. Of note, OctoMed-7B was released in November 2025 and is a reasoning model.[21] These findings are consistent with the general momentum of the LLM field, which is seeking methods to develop smaller LLMs that facilitate deployment without large computational resources though with similar or superior performance than larger LLMs.[22]

Most of the LLMs tested had fewer than 10 billion parameters, resulting in moderate to low performance in this task. This outcome was expected given the limited number of parameters, compared to larger models such as medgemma-27b-text-it and Llama3-OpenBioLLM-70B,[9,13-15] or recent LLMs of similar size yet with greater reasoning capabilities such as OctoMed-7B.[21] Despite the initial performance of the smallest LLM (medgemma-4b-it), further fine-tuning with task-specific data led to an improvement in its capabilities. This enhancement surpassed the performance of other LLMs with fewer than 10 billion parameters and even rivaled the largest LLM (70B). This outcome may suggest that the LLMs lacked comprehensive knowledge about the epidemiology of Peru and the typical formulation of medical questions in Peru.[13] By fine-tuning medgemma-4b-it, it utilized its extensive underlying knowledge base and acquired proficiency in the medical field of Peru. Thus, this fine-tuning resulted in substantial performance improvements.

**Contributions**

We have released PeruMedQA, providing a comprehensive description of its construction process. To the best of our knowledge, despite the examinations conducted by CONAREME being open access and available on their website,[16] this is the first instance where these questions have been harmonized and systematically organized in a computer-readable format. We make this dataset open access through this publication, enabling others to verify our findings and develop AI systems. We also envision this dataset as a valuable resource for medical professionals preparing for upcoming CONAREME examinations.

There is growing interest in enhancing the capabilities of LLMs in various clinical scenarios, particularly in addressing clinical tasks across specialties and for diverse populations. For instance, to achieve this objective, OpenAI launched HealthBench in April 2025, a data resource designed to assess the capabilities of AI systems in the healthcare domain.[23] The initial step in improving the capabilities of LLMs in diverse medical scenarios, is to empirically evaluate the performance of existing medical LLMs in comprehending clinical knowledge from diverse populations. Although similar investigations have been conducted with medical examinations from Brazil and the African continent,[14,15] an analysis of medical examinations in Spanish from a South American country like Peru, characterized by a range of epidemiological profiles including infectious diseases, tropical diseases, neglected diseases, and chronic noncommunicable diseases,[24] has not yet been undertaken. In general, the same argument can be applied to countries located in the southern hemisphere. Researchers and AI developers can utilize our findings as a benchmark and employ the LLMs that showed the most promising results in downstream applications.

As the smallest model we evaluated, medgemma-4b-it would be the most resource-friendly without requiring extensive computational resources. Consequently, we aimed to enhance its performance in this task by fine-tuning medgemma-4b-it to deliver a resource-efficient LLM with superior accuracy compared to the original medgemma-4b-it.[13,19] As the fine-tuned medgemma-4b-it demonstrated superior performance compared to the vanilla medgemma-4b-it and other LLMs we evaluated, and as the fine-tuned version of medgemma-4b-it rivaled a 70B parameter, the medgemma-4b-it-FT would emerge as a valuable medical LLM capable of addressing medical inquiries in Spanish from Peru and potentially applicable to other nations with comparable epidemiological characteristics. Nevertheless, it is worth remembering that while fine-tuning on domain-specific data could improve performance, it requires careful evaluation of catastrophic forgetting, where domain gains may degrade previously learned knowledge, thus requiring benchmarking across diverse contexts to balance specialization and generalization.

## Strengths and limitations

This systematic evaluation of medical LLMs on exams for medical professionals pursuing further training in Peru, utilizing Spanish language questions (PeruMedQA), is a novel contribution. Ten LLMs were tested with a consistent methodology, and a fine-tuned version was developed, achieving superior performance compared to the vanilla version.

However, this study has limitations. First, the evaluation was limited to open-access LLMs, and we excluded general-purpose LLMs. Recent state-of-the-art LLMs, such as GPT-5, could potentially yield better results. Nevertheless, in the medical domain, researchers and developers may prefer LLMs specifically targeting this field. The focus on open-access LLMs ensures that others can directly benefit from our findings. Second, given the substantial resources required, we did not fine-tune medgemma-27b-text-it, a model that could potentially achieve perfect results. Moreover, even if a fine-tuned version of medgemma-27b-text-it had been developed, fewer researchers may have access to the computational resources needed to load and make inference using this model. The medgemma-4b-it, including our fine-tuned version, can be utilized in platforms such as Google Collaboratory and some user-graded computers with GPUs. Third, we did not explore the rationale or justification behind the answers provided by the LLMs. Our primary objective was to verify whether LLMs can answer medical questions in Spanish from Peru and assess their accuracy. Future work should prompt the LLMs to provide justifications or elaborate on the reasons behind their choices and evaluate whether these elaborations are accurate and meaningful. Additionally, the evaluation methodology could be enhanced by allowing smaller LLMs to generate free-form answers and then using a larger, more capable LLM as a judge to extract the final answer from the raw text, thereby avoiding strict formatting requirements that demand strong instruction-following capabilities which smaller models may lack. Fourth, we curated and employed a dataset of medical questions for medical professionals embarking on specialty and subspecialty training. This dataset excluded questions pertinent to medical students (e.g., the national examination prior to medical licensing) and other healthcare professionals. The extent to which the LLMs we evaluated, or other LLMs, can provide accurate responses to those examinations remains unexplored and warrants empirical verification. Fifth, given the extensive number of questions we curated, we did not distinguish between questions that inquired about a concept or a fact, and questions that proposed a clinical case. Future research should examine questions that propose a clinical case to assess whether LLMs adhere to a logical reasoning to arrive at the final question. Sixth, to maintain consistency with other comparable evaluations, we utilized a zero-shot prompt.[13-15] Although other advanced prompting techniques, such as chain of thought or three of thoughts, could have yielded superior outcomes, a comparison of prompt techniques to attain higher accuracies will be subject of future work. Seventh, to ensure consistency in our dataset (PeruMedQA) regarding the number of multiple-choice answers, in some years we introduced an additional alternative to reach a total of five choices for each question. This addition of potential answers subtly increases the difficulty of the questions, as each choice transitions from a probability of 1/4 to 1/5. Consequently, direct head-to-head comparisons with published grades of these exams achieved by humans were not feasible.

## Conclusions

Although not primarily trained on medical data in Spanish, nor from the Peruvian medical context or epidemiological profile, medical LLMs have shown the ability to successfully answer advanced medical questions for medical professionals in Peru. Notably, medgemma-27b-text-it exhibited exceptional performance followed by OctoMed-7B, and Llama3-OpenBioLLM-70B as well as a fine-tuned version of medgemma-4b-it that leveraged PEFT and LoRA. For medical AI applications and research that necessitate knowledge bases from Spanish-speaking countries and those exhibiting similar epidemiological profiles to Peru's, researchers and developers should utilize medgemma-27b-text-it, OctoMed-7B, or the fine-tuned version of medgemma-4b-it. The fine-tuned version of medgemma-4b-it demonstrated good performance without the requirement for substantial computational resources, unlike LLMs with 27 or 70 billion parameters.

## REFERENCES

1. Meng X, Yan X, Zhang K, et al. The application of large language models in medicine: A scoping review. *iScience* 2024; **27**(5): 109713.

2. Shool S, Adimi S, Saboori Amleshi R, Bitaraf E, Golpira R, Tara M. A systematic review of large language model (LLM) evaluations in clinical medicine. *BMC Medical Informatics and Decision Making* 2025; **25**(1): 117.

3. Yu E, Chu X, Zhang W, et al. Large Language Models in Medicine: Applications, Challenges, and Future Directions. *Int J Med Sci* 2025; **22**(11): 2792-801.

4. Jung KH. Large Language Models in Medicine: Clinical Applications, Technical Challenges, and Ethical Considerations. *Healthc Inform Res* 2025; **31**(2): 114-24.

5. Maity S, Saikia MJ. Large Language Models in Healthcare and Medical Applications: A Review. *Bioengineering (Basel)* 2025; **12**(6).

6. Singhal K, Tu T, Gottweis J, et al. Toward expert-level medical question answering with large language models. *Nature Medicine* 2025; **31**(3): 943-50.

7. Noda R, Tanabe K, Ichikawa D, Shibagaki Y. GPT-4's performance in supporting physician decision-making in nephrology multiple-choice questions. *Sci Rep* 2025; **15**(1): 15439.

8. Miranda J, Pereira-Silva R, Guichard J, Meneses J, Carreira AN, Seixas D. Artificial Intelligence Outperforms Physicians in General Medical Knowledge, Except in the Paediatrics Domain: A Cross-Sectional Study. *Bioengineering (Basel)* 2025; **12**(6).

9. Abrantes J. Assessing Large Language Models for Medical Question Answering in Portuguese: Open-Source Versus Closed-Source Approaches. *Cureus* 2025; **17**(5): e84165.

10. Riina N, Patlolla L, Hernandez Joya C, Bautista R, Olivar-Villanueva M, Kumar A. An Evaluation of English to Spanish Medical Translation by Large Language Models. 2024 September; Chicago, USA: Association for Machine Translation in the Americas; 2024. p. 222-36.

11. GBD 2021 Demographics Collaborators. Global age-sex-specific mortality, life expectancy, and population estimates in 204 countries and territories and 811 subnational locations, 1950-2021, and the impact of the COVID-19 pandemic: a comprehensive demographic analysis for the Global Burden of Disease Study 2021. *Lancet* 2024.

12. GBD 2021 Diseases and Injuries Collaborators. Global incidence, prevalence, years lived with disability (YLDs), disability-adjusted life-years (DALYs), and healthy life expectancy (HALE) for 371 diseases and injuries in 204 countries and territories and 811 subnational locations, 1990–2021: a systematic analysis for the Global Burden of Disease Study 2021. *The Lancet*.

13. Sellergren A, Kazemzadeh S, Jaroensri T, et al. Medgemma technical report. *arXiv preprint arXiv:250705201* 2025.

14. D'addario AMV. HealthQA-BR: A System-Wide Benchmark Reveals Critical Knowledge Gaps in Large Language Models. *arXiv preprint arXiv:250621578* 2025.

15. Olatunji T, Nimo C, Owodunni A, et al. AfriMed-QA: a Pan-African, multi-specialty, medical question-answering benchmark dataset. *arXiv preprint arXiv:241115640* 2024.

16. CONAREME Consejo Nacional de Residentado Medico [Internet]. [cited 6 September 2025]. URL: https://www.conareme.org.pe/web/.

17. Scraping PDF documents containing exams with highlighted correct answers [Internet]. [cited 6 September 2025]. URL: https://medium.com/gitconnected/scraping-pdf-documents-containing-exams-with-highlighted-correct-answers-68d0a6e9b397.

18. Hendrycks D, Burns C, Basart S, et al. Measuring massive multitask language understanding. *arXiv preprint arXiv:200903300* 2020.

19. Google-Health/medgemma [Internet]. [cited 6 September 2025]. URL: https://github.com/google-health/medgemma/blob/main/notebooks/fine_tune_with_hugging_face.ipynb.

20. Chung HW, Hou L, Longpre S, et al. Scaling instruction-finetuned language models. *Journal of Machine Learning Research* 2024; **25**(70): 1-53.

21. Ossowski T, Zhang S, Liu Q, et al. OctoMed: Data Recipes for State-of-the-Art Multimodal Medical Reasoning. *arXiv preprint arXiv:251123269* 2025.

22. Kim H, Hwang H, Lee J, et al. Small language models learn enhanced reasoning skills from medical textbooks. *NPJ Digit Med* 2025; **8**(1): 240.

23. Arora RK, Wei J, Hicks RS, et al. Healthbench: Evaluating large language models towards improved human health. *arXiv preprint arXiv:250508775* 2025.

24. Carrillo-Larco RM, Guzman-Vilca WC, Leon-Velarde F, et al. Peru - Progress in health and sciences in 200 years of independence. *Lancet Reg Health Am* 2022; **7**: 100148.

## DISCLOSURES

**Acknowledgements**: All the models tested herein were executed on the HyPER C3 – Community Cloud HPC Cluster at Emory University, Atlanta, USA. https://it.emory.edu/catalog/cloud-services/hyper-c3.html The research leading to these results has been funded by grant PID2023-150515OB-I00 from the Spanish Government, partially covered with funds from the Recovery and Resilience Facility (RRF).

**Conflict of interests:** None.

**Data sharing:** We developed PeruMedQA, which we will release as an open-access resource. PeruMedQA is hosted in a GitHub repository (https://github.com/rodrigo-carrillo/PeruMedQA). This resource has apache-2 license, allowing commercial and non-commercial use.

**Code sharing:** The analysis code (Jupyter Notebooks) for all the analytical steps is stored in a GitHub repository (https://github.com/rodrigo-carrillo/PeruMedQA). These resources have apache-2 license, allowing commercial and non-commercial use.

## SUPPLEMENTARY MATERIALS

### Supplementary Table 1. Number of questions per specialty and year.

| | Name (Spanish) | Name (English) | Years | | | | | | | |
|---|---|---|---|---|---|---|---|---|---|---|
| | | | 2018 | 2019 | 2020 | 2022 | 2023 | 2024 | 2025 | Total |
| subspecialty | Anatomía Patológica y Patología | Pathology and Anatomical Pathology | 100 | 100 | 100 | 0 | 100 | 100 | 100 | 600 |
| | Anestesiología | Anesthesiology | 100 | 100 | 100 | 100 | 100 | 100 | 100 | 700 |
| | Cirugia general | General Surgery | 100 | 100 | 100 | 100 | 100 | 100 | 100 | 700 |
| | Cirugia de Torax y Cardiovascular | Thoracic and Cardiovascular Surgery | 0 | 100 | 0 | 100 | 100 | 100 | 200 | 600 |
| | Ginecología & Obstetricia | Gynecology & Obstetrics | 100 | 100 | 100 | 100 | 100 | 100 | 100 | 700 |
| | Neurocirugía | Neurosurgery | 0 | 0 | 0 | 100 | 100 | 100 | 100 | 400 |
| | Oftalmología | Ophthalmology | 100 | 100 | 100 | 100 | 100 | 100 | 100 | 700 |
| | Pediatría | Pediatrics | 100 | 100 | 100 | 100 | 100 | 100 | 100 | 700 |
| | Psiquiatría | Psychiatry | 100 | 100 | 100 | 100 | 100 | 100 | 100 | 700 |
| | Radiología | Radiology | 100 | 100 | 0 | 100 | 100 | 100 | 100 | 600 |
| | Urología | Urology | 100 | 0 | 100 | 100 | 100 | 100 | 100 | 600 |
| specialty* | Prueba A | Test A | 100 | 100 | 90 | 100 | 100 | 100 | 100 | 690 |
| | Prueba B | Test B | 100 | 100 | 90 | 100 | 100 | 100 | 100 | 690 |
| | | | | | | | | | Total | 8380 |

*For the specialty examination, there are two distinct, yet equivalent in complexity, versions of the test (Test A and Test B).

**Supplementary Table 2. Zero-shot prompt.**

| System message | "Eres un asistente médico experto con entrenamiento en Perú." |
|---|---|
| User message | user_prompt = """<br>Instrucciones: Las siguientes son preguntas de opción múltiple sobre conocimientos médicos.<br>Resuélvalas paso a paso, comenzando por resumir *internamente* la información disponible y termine con "Respuesta final:" seguido *solo* de la letra correspondiente a la respuesta correcta.<br>Por ejemplo: 'Respuesta final:X'.<br>No incluya en su respuesta el razonamiento paso a paso que hizo internamente.<br>Escriba una sola opción de las cinco como respuesta final.<br>Pregunta: " {original_question} "<br>""" |

In the user message, we incorporated the question, along with the five multiple-choice options, in the following format: *Varón de 65 años con antecedente de hipertensión arterial, acude por cefalea intensa, convulsiones y edema de papila. Examen: PA: 190/120 mmHg. ¿Cuál es el medicamento inicial a usar? Varón de 65 años con antecedente de hipertensión arterial, acude por cefalea intensa, convulsiones y edema de papila. Examen: PA: 190/120 mmHg. ¿Cuál es el medicamento inicial a usar? A) Labetalol B) Hidralazina C) Nitroprusiato D) Nitroglicerina E) Ninguna de las anteriores*

**Supplementary Table 3. Ten large language models (LLMs) utilized for medical question answering.**

| Model | Parameters | Model Identifier | Website |
|---|---|---|---|
| medgemma-4b-it | 4B | google/medgemma-4b-it | https://huggingface.co/google/medgemma-4b-it |
| BioMistral-7B-DARE | 7B | BioMistral/BioMistral-7B-DARE | https://huggingface.co/BioMistral/BioMistral-7B-DARE |
| MediPhi-Instruct-3.8B | 3.8B | microsoft/MediPhi-Instruct | https://huggingface.co/microsoft/MediPhi-Instruct |
| Llama3-OpenBioLLM-8B | 8B | aaditya/Llama3-OpenBioLLM-8B | https://huggingface.co/aaditya/Llama3-OpenBioLLM-8B |
| JSL-MedLlama-3-8B-v2.0 | 8B | johnsnowlabs/JSL-MedLlama-3-8B-v2.0 | https://huggingface.co/johnsnowlabs/JSL-MedLlama-3-8B-v2.0 |
| meditron-7b | 7B | epfl-llm/meditron-7b | https://huggingface.co/epfl-llm/meditron-7b |
| OctoMed-7B | 7B | OctoMed/OctoMed-7B | https://huggingface.co/OctoMed/OctoMed-7B |
| medgemma-1.5-4b-it | 4B | google/medgemma-1.5-4b-it | https://huggingface.co/google/medgemma-1.5-4b-it |
| medgemma-27b-text-it | 27B | google/medgemma-27b-text-it | https://huggingface.co/google/medgemma-27b-text-it |
| Llama3-OpenBioLLM-70B | 70B | aaditya/Llama3-OpenBioLLM-70B | https://huggingface.co/aaditya/Llama3-OpenBioLLM-70B |

The number of parameters is expressed in the billions (B) format.

**Supplementary Table 4. Number of questions without a valid answer.**

| Model | Number of questions without valid answer by LLM (hallucinations). |
|---|---|
| medgemma-4b-it | 12 / 8,380 (0.14%) |
| BioMistral-7B-DARE | 76 / 8,380 (0.90%) |
| MediPhi-Instruct-3.8B | 4 / 8,380 (0.04%) |
| Llama3-OpenBioLLM-8B | 416 / 8,380 (4.96%) |
| JSL-MedLlama-3-8B-v2.0 | 755 / 8,380 (9.00%) |
| meditron-7b | 5,562 / 8,380 (66.37%) |
| OctoMed-7B | 15 / 8,380 (0.18%) |
| medgemma-1.5-4b-it | 231 / 8,380 (2.76%) |
| medgemma-27b-text-it | 2 / 8,380 (0.02%) |
| Llama3-OpenBioLLM-70B | 0 / 8,380 (0.00%) |
| medgemma-4b-it fine-tuned | 0 / 8,380 (0.00%) |

The medgemma-4b-it fine-tuned (last row) refers to the model we fine-tuned using parameter efficient fine tuning (PEFT) with Low-Rank Adaptation (LoRA) with the base model being medgemma-4b-it.

## Supplementary Table 5. Underlying results by model, exam and year.

| Model | Exam | Year | Correct answers (N) | Total questions | Percent correct answers |
|---|---|---|---|---|---|
| meditron-7b | Pathological Anatomy and Pathology | 2018 | 6 | 31 | 19.35 |
| meditron-7b | Pathological Anatomy and Pathology | 2019 | 6 | 30 | 20.00 |
| meditron-7b | Pathological Anatomy and Pathology | 2020 | 10 | 50 | 20.00 |
| meditron-7b | Pathological Anatomy and Pathology | 2023 | 5 | 29 | 17.24 |
| meditron-7b | Pathological Anatomy and Pathology | 2024 | 7 | 37 | 18.92 |
| meditron-7b | Pathological Anatomy and Pathology | 2025 | 12 | 54 | 22.22 |
| meditron-7b | Anesthesiology | 2018 | 2 | 15 | 13.33 |
| meditron-7b | Anesthesiology | 2019 | 7 | 33 | 21.21 |
| meditron-7b | Anesthesiology | 2020 | 3 | 24 | 12.50 |
| meditron-7b | Anesthesiology | 2022 | 7 | 21 | 33.33 |
| meditron-7b | Anesthesiology | 2023 | 11 | 30 | 36.67 |
| meditron-7b | Anesthesiology | 2024 | 4 | 19 | 21.05 |
| meditron-7b | Anesthesiology | 2025 | 3 | 25 | 12.00 |
| meditron-7b | General Surgery | 2018 | 8 | 31 | 25.81 |
| meditron-7b | General Surgery | 2019 | 6 | 20 | 30.00 |
| meditron-7b | General Surgery | 2020 | 8 | 36 | 22.22 |
| meditron-7b | General Surgery | 2022 | 6 | 34 | 17.65 |
| meditron-7b | General Surgery | 2023 | 12 | 33 | 36.36 |
| meditron-7b | General Surgery | 2024 | 8 | 45 | 17.78 |
| meditron-7b | General Surgery | 2025 | 11 | 30 | 36.67 |
| meditron-7b | Thoracic and Cardiovascular Surgery | 2019 | 12 | 46 | 26.09 |
| meditron-7b | Thoracic and Cardiovascular Surgery | 2022 | 7 | 43 | 16.28 |
| meditron-7b | Thoracic and Cardiovascular Surgery | 2023 | 8 | 37 | 21.62 |
| meditron-7b | Thoracic and Cardiovascular Surgery | 2024 | 9 | 35 | 25.71 |
| meditron-7b | Thoracic and Cardiovascular Surgery | 2025 | 26 | 64 | 40.62 |
| meditron-7b | Gynecology and Obstetrics | 2018 | 9 | 43 | 20.93 |
| meditron-7b | Gynecology and Obstetrics | 2019 | 8 | 29 | 27.59 |
| meditron-7b | Gynecology and Obstetrics | 2020 | 12 | 41 | 29.27 |
| meditron-7b | Gynecology and Obstetrics | 2022 | 5 | 36 | 13.89 |
| meditron-7b | Gynecology and Obstetrics | 2023 | 11 | 40 | 27.50 |
| meditron-7b | Gynecology and Obstetrics | 2024 | 6 | 40 | 15.00 |
| meditron-7b | Gynecology and Obstetrics | 2025 | 5 | 35 | 14.29 |
| meditron-7b | Neurosurgery | 2022 | 8 | 34 | 23.53 |

| meditron-7b | Neurosurgery | 2023 | 8 | 24 | 33.33 |
|---|---|---|---|---|---|
| meditron-7b | Neurosurgery | 2024 | 6 | 37 | 16.22 |
| meditron-7b | Neurosurgery | 2025 | 11 | 45 | 24.44 |
| meditron-7b | Ophthalmology | 2018 | 8 | 38 | 21.05 |
| meditron-7b | Ophthalmology | 2019 | 5 | 29 | 17.24 |
| meditron-7b | Ophthalmology | 2020 | 8 | 34 | 23.53 |
| meditron-7b | Ophthalmology | 2022 | 7 | 30 | 23.33 |
| meditron-7b | Ophthalmology | 2023 | 7 | 29 | 24.14 |
| meditron-7b | Ophthalmology | 2024 | 6 | 27 | 22.22 |
| meditron-7b | Ophthalmology | 2025 | 9 | 39 | 23.08 |
| meditron-7b | Pediatrics | 2018 | 5 | 33 | 15.15 |
| meditron-7b | Pediatrics | 2019 | 11 | 41 | 26.83 |
| meditron-7b | Pediatrics | 2020 | 15 | 50 | 30.00 |
| meditron-7b | Pediatrics | 2022 | 8 | 45 | 17.78 |
| meditron-7b | Pediatrics | 2023 | 14 | 48 | 29.17 |
| meditron-7b | Pediatrics | 2024 | 5 | 30 | 16.67 |
| meditron-7b | Pediatrics | 2025 | 7 | 40 | 17.50 |
| meditron-7b | Test A | 2018 | 4 | 28 | 14.29 |
| meditron-7b | Test A | 2019 | 7 | 36 | 19.44 |
| meditron-7b | Test A | 2020 | 6 | 35 | 17.14 |
| meditron-7b | Test A | 2022 | 10 | 36 | 27.78 |
| meditron-7b | Test A | 2023 | 6 | 27 | 22.22 |
| meditron-7b | Test A | 2024 | 7 | 28 | 25.00 |
| meditron-7b | Test A | 2025 | 9 | 34 | 26.47 |
| meditron-7b | Test B | 2018 | 9 | 36 | 25.00 |
| meditron-7b | Test B | 2019 | 12 | 32 | 37.50 |
| meditron-7b | Test B | 2020 | 10 | 34 | 29.41 |
| meditron-7b | Test B | 2022 | 8 | 32 | 25.00 |
| meditron-7b | Test B | 2023 | 5 | 32 | 15.62 |
| meditron-7b | Test B | 2024 | 11 | 39 | 28.21 |
| meditron-7b | Test B | 2025 | 9 | 31 | 29.03 |
| meditron-7b | Psychiatry | 2018 | 5 | 22 | 22.73 |
| meditron-7b | Psychiatry | 2019 | 6 | 28 | 21.43 |
| meditron-7b | Psychiatry | 2020 | 7 | 18 | 38.89 |
| meditron-7b | Psychiatry | 2022 | 7 | 27 | 25.93 |
| meditron-7b | Psychiatry | 2023 | 6 | 30 | 20.00 |
| meditron-7b | Psychiatry | 2024 | 5 | 31 | 16.13 |
| meditron-7b | Psychiatry | 2025 | 5 | 17 | 29.41 |
| meditron-7b | Radiology | 2018 | 9 | 24 | 37.50 |

| meditron-7b | Radiology | 2019 | 7 | 36 | 19.44 |
|---|---|---|---|---|---|
| meditron-7b | Radiology | 2022 | 6 | 30 | 20.00 |
| meditron-7b | Radiology | 2023 | 6 | 37 | 16.22 |
| meditron-7b | Radiology | 2024 | 11 | 40 | 27.50 |
| meditron-7b | Radiology | 2025 | 5 | 39 | 12.82 |
| meditron-7b | Urology | 2018 | 11 | 42 | 26.19 |
| meditron-7b | Urology | 2020 | 7 | 39 | 17.95 |
| meditron-7b | Urology | 2022 | 4 | 31 | 12.90 |
| meditron-7b | Urology | 2023 | 9 | 30 | 30.00 |
| meditron-7b | Urology | 2024 | 9 | 29 | 31.03 |
| meditron-7b | Urology | 2025 | 7 | 39 | 17.95 |
| Llama3-OpenBioLLM-8B | Pathological Anatomy and Pathology | 2018 | 35 | 91 | 38.46 |
| Llama3-OpenBioLLM-8B | Pathological Anatomy and Pathology | 2019 | 32 | 92 | 34.78 |
| Llama3-OpenBioLLM-8B | Pathological Anatomy and Pathology | 2020 | 33 | 97 | 34.02 |
| Llama3-OpenBioLLM-8B | Pathological Anatomy and Pathology | 2023 | 32 | 97 | 32.99 |
| Llama3-OpenBioLLM-8B | Pathological Anatomy and Pathology | 2024 | 23 | 98 | 23.47 |
| Llama3-OpenBioLLM-8B | Pathological Anatomy and Pathology | 2025 | 24 | 98 | 24.49 |
| Llama3-OpenBioLLM-8B | Anesthesiology | 2018 | 27 | 94 | 28.72 |
| Llama3-OpenBioLLM-8B | Anesthesiology | 2019 | 31 | 92 | 33.70 |
| Llama3-OpenBioLLM-8B | Anesthesiology | 2020 | 33 | 88 | 37.50 |
| Llama3-OpenBioLLM-8B | Anesthesiology | 2022 | 40 | 95 | 42.11 |
| Llama3-OpenBioLLM-8B | Anesthesiology | 2023 | 21 | 92 | 22.83 |
| Llama3-OpenBioLLM-8B | Anesthesiology | 2024 | 25 | 97 | 25.77 |
| Llama3-OpenBioLLM-8B | Anesthesiology | 2025 | 31 | 95 | 32.63 |
| Llama3-OpenBioLLM-8B | General Surgery | 2018 | 30 | 96 | 31.25 |
| Llama3-OpenBioLLM-8B | General Surgery | 2019 | 32 | 96 | 33.33 |
| Llama3-OpenBioLLM-8B | General Surgery | 2020 | 31 | 98 | 31.63 |
| Llama3-OpenBioLLM-8B | General Surgery | 2022 | 34 | 97 | 35.05 |
| Llama3-OpenBioLLM-8B | General Surgery | 2023 | 27 | 93 | 29.03 |
| Llama3-OpenBioLLM-8B | General Surgery | 2024 | 20 | 96 | 20.83 |
| Llama3-OpenBioLLM-8B | General Surgery | 2025 | 24 | 100 | 24.00 |
| Llama3-OpenBioLLM-8B | Thoracic and Cardiovascular Surgery | 2019 | 28 | 92 | 30.43 |
| Llama3-OpenBioLLM-8B | Thoracic and Cardiovascular Surgery | 2022 | 35 | 93 | 37.63 |
| Llama3-OpenBioLLM-8B | Thoracic and Cardiovascular Surgery | 2023 | 20 | 97 | 20.62 |
| Llama3-OpenBioLLM-8B | Thoracic and Cardiovascular Surgery | 2024 | 19 | 93 | 20.43 |
| Llama3-OpenBioLLM-8B | Thoracic and Cardiovascular Surgery | 2025 | 30 | 188 | 15.96 |

| Llama3-OpenBioLLM-8B | Gynecology and Obstetrics | 2018 | 29 | 91 | 31.87 |
|---|---|---|---|---|---|
| Llama3-OpenBioLLM-8B | Gynecology and Obstetrics | 2019 | 33 | 93 | 35.48 |
| Llama3-OpenBioLLM-8B | Gynecology and Obstetrics | 2020 | 34 | 99 | 34.34 |
| Llama3-OpenBioLLM-8B | Gynecology and Obstetrics | 2022 | 36 | 95 | 37.89 |
| Llama3-OpenBioLLM-8B | Gynecology and Obstetrics | 2023 | 18 | 91 | 19.78 |
| Llama3-OpenBioLLM-8B | Gynecology and Obstetrics | 2024 | 23 | 95 | 24.21 |
| Llama3-OpenBioLLM-8B | Gynecology and Obstetrics | 2025 | 21 | 92 | 22.83 |
| Llama3-OpenBioLLM-8B | Neurosurgery | 2022 | 31 | 99 | 31.31 |
| Llama3-OpenBioLLM-8B | Neurosurgery | 2023 | 15 | 97 | 15.46 |
| Llama3-OpenBioLLM-8B | Neurosurgery | 2024 | 28 | 92 | 30.43 |
| Llama3-OpenBioLLM-8B | Neurosurgery | 2025 | 19 | 95 | 20.00 |
| Llama3-OpenBioLLM-8B | Ophthalmology | 2018 | 23 | 90 | 25.56 |
| Llama3-OpenBioLLM-8B | Ophthalmology | 2019 | 28 | 91 | 30.77 |
| Llama3-OpenBioLLM-8B | Ophthalmology | 2020 | 32 | 94 | 34.04 |
| Llama3-OpenBioLLM-8B | Ophthalmology | 2022 | 33 | 96 | 34.38 |
| Llama3-OpenBioLLM-8B | Ophthalmology | 2023 | 18 | 98 | 18.37 |
| Llama3-OpenBioLLM-8B | Ophthalmology | 2024 | 15 | 91 | 16.48 |
| Llama3-OpenBioLLM-8B | Ophthalmology | 2025 | 21 | 95 | 22.11 |
| Llama3-OpenBioLLM-8B | Pediatrics | 2018 | 32 | 96 | 33.33 |
| Llama3-OpenBioLLM-8B | Pediatrics | 2019 | 40 | 96 | 41.67 |
| Llama3-OpenBioLLM-8B | Pediatrics | 2020 | 33 | 94 | 35.11 |
| Llama3-OpenBioLLM-8B | Pediatrics | 2022 | 35 | 97 | 36.08 |
| Llama3-OpenBioLLM-8B | Pediatrics | 2023 | 28 | 95 | 29.47 |
| Llama3-OpenBioLLM-8B | Pediatrics | 2024 | 28 | 99 | 28.28 |
| Llama3-OpenBioLLM-8B | Pediatrics | 2025 | 24 | 94 | 25.53 |
| Llama3-OpenBioLLM-8B | Test A | 2018 | 38 | 95 | 40.00 |
| Llama3-OpenBioLLM-8B | Test A | 2019 | 41 | 95 | 43.16 |
| Llama3-OpenBioLLM-8B | Test A | 2020 | 44 | 87 | 50.57 |
| Llama3-OpenBioLLM-8B | Test A | 2022 | 35 | 98 | 35.71 |
| Llama3-OpenBioLLM-8B | Test A | 2023 | 33 | 94 | 35.11 |
| Llama3-OpenBioLLM-8B | Test A | 2024 | 28 | 99 | 28.28 |
| Llama3-OpenBioLLM-8B | Test A | 2025 | 33 | 97 | 34.02 |
| Llama3-OpenBioLLM-8B | Test B | 2018 | 42 | 98 | 42.86 |
| Llama3-OpenBioLLM-8B | Test B | 2019 | 35 | 92 | 38.04 |
| Llama3-OpenBioLLM-8B | Test B | 2020 | 26 | 88 | 29.55 |
| Llama3-OpenBioLLM-8B | Test B | 2022 | 36 | 94 | 38.30 |
| Llama3-OpenBioLLM-8B | Test B | 2023 | 27 | 96 | 28.12 |
| Llama3-OpenBioLLM-8B | Test B | 2024 | 20 | 96 | 20.83 |
| Llama3-OpenBioLLM-8B | Test B | 2025 | 25 | 96 | 26.04 |

| Llama3-OpenBioLLM-8B | Psychiatry | 2018 | 37 | 97 | 38.14 |
|---|---|---|---|---|---|
| Llama3-OpenBioLLM-8B | Psychiatry | 2019 | 55 | 95 | 57.89 |
| Llama3-OpenBioLLM-8B | Psychiatry | 2020 | 51 | 95 | 53.68 |
| Llama3-OpenBioLLM-8B | Psychiatry | 2022 | 47 | 97 | 48.45 |
| Llama3-OpenBioLLM-8B | Psychiatry | 2023 | 45 | 98 | 45.92 |
| Llama3-OpenBioLLM-8B | Psychiatry | 2024 | 48 | 94 | 51.06 |
| Llama3-OpenBioLLM-8B | Psychiatry | 2025 | 42 | 98 | 42.86 |
| Llama3-OpenBioLLM-8B | Radiology | 2018 | 33 | 97 | 34.02 |
| Llama3-OpenBioLLM-8B | Radiology | 2019 | 29 | 91 | 31.87 |
| Llama3-OpenBioLLM-8B | Radiology | 2022 | 27 | 97 | 27.84 |
| Llama3-OpenBioLLM-8B | Radiology | 2023 | 26 | 99 | 26.26 |
| Llama3-OpenBioLLM-8B | Radiology | 2024 | 20 | 99 | 20.20 |
| Llama3-OpenBioLLM-8B | Radiology | 2025 | 26 | 98 | 26.53 |
| Llama3-OpenBioLLM-8B | Urology | 2018 | 44 | 95 | 46.32 |
| Llama3-OpenBioLLM-8B | Urology | 2020 | 35 | 93 | 37.63 |
| Llama3-OpenBioLLM-8B | Urology | 2022 | 32 | 92 | 34.78 |
| Llama3-OpenBioLLM-8B | Urology | 2023 | 20 | 91 | 21.98 |
| Llama3-OpenBioLLM-8B | Urology | 2024 | 23 | 90 | 25.56 |
| Llama3-OpenBioLLM-8B | Urology | 2025 | 18 | 93 | 19.35 |
| BioMistral-7B-DARE | Pathological Anatomy and Pathology | 2018 | 36 | 100 | 36.00 |
| BioMistral-7B-DARE | Pathological Anatomy and Pathology | 2019 | 35 | 100 | 35.00 |
| BioMistral-7B-DARE | Pathological Anatomy and Pathology | 2020 | 28 | 100 | 28.00 |
| BioMistral-7B-DARE | Pathological Anatomy and Pathology | 2023 | 37 | 99 | 37.37 |
| BioMistral-7B-DARE | Pathological Anatomy and Pathology | 2024 | 37 | 97 | 38.14 |
| BioMistral-7B-DARE | Pathological Anatomy and Pathology | 2025 | 27 | 98 | 27.55 |
| BioMistral-7B-DARE | Anesthesiology | 2018 | 20 | 98 | 20.41 |
| BioMistral-7B-DARE | Anesthesiology | 2019 | 27 | 97 | 27.84 |
| BioMistral-7B-DARE | Anesthesiology | 2020 | 32 | 98 | 32.65 |
| BioMistral-7B-DARE | Anesthesiology | 2022 | 28 | 100 | 28.00 |
| BioMistral-7B-DARE | Anesthesiology | 2023 | 40 | 100 | 40.00 |
| BioMistral-7B-DARE | Anesthesiology | 2024 | 28 | 99 | 28.28 |
| BioMistral-7B-DARE | Anesthesiology | 2025 | 38 | 98 | 38.78 |
| BioMistral-7B-DARE | General Surgery | 2018 | 25 | 100 | 25.00 |
| BioMistral-7B-DARE | General Surgery | 2019 | 32 | 100 | 32.00 |
| BioMistral-7B-DARE | General Surgery | 2020 | 31 | 99 | 31.31 |
| BioMistral-7B-DARE | General Surgery | 2022 | 27 | 100 | 27.00 |
| BioMistral-7B-DARE | General Surgery | 2023 | 25 | 100 | 25.00 |

| BioMistral-7B-DARE | General Surgery | 2024 | 29 | 100 | 29.00 |
|---|---|---|---|---|---|
| BioMistral-7B-DARE | General Surgery | 2025 | 31 | 100 | 31.00 |
| BioMistral-7B-DARE | Thoracic and Cardiovascular Surgery | 2019 | 30 | 98 | 30.61 |
| BioMistral-7B-DARE | Thoracic and Cardiovascular Surgery | 2022 | 32 | 97 | 32.99 |
| BioMistral-7B-DARE | Thoracic and Cardiovascular Surgery | 2023 | 43 | 100 | 43.00 |
| BioMistral-7B-DARE | Thoracic and Cardiovascular Surgery | 2024 | 34 | 96 | 35.42 |
| BioMistral-7B-DARE | Thoracic and Cardiovascular Surgery | 2025 | 70 | 196 | 35.71 |
| BioMistral-7B-DARE | Gynecology and Obstetrics | 2018 | 20 | 99 | 20.20 |
| BioMistral-7B-DARE | Gynecology and Obstetrics | 2019 | 27 | 99 | 27.27 |
| BioMistral-7B-DARE | Gynecology and Obstetrics | 2020 | 30 | 99 | 30.30 |
| BioMistral-7B-DARE | Gynecology and Obstetrics | 2022 | 34 | 100 | 34.00 |
| BioMistral-7B-DARE | Gynecology and Obstetrics | 2023 | 31 | 100 | 31.00 |
| BioMistral-7B-DARE | Gynecology and Obstetrics | 2024 | 31 | 99 | 31.31 |
| BioMistral-7B-DARE | Gynecology and Obstetrics | 2025 | 34 | 99 | 34.34 |
| BioMistral-7B-DARE | Neurosurgery | 2022 | 22 | 98 | 22.45 |
| BioMistral-7B-DARE | Neurosurgery | 2023 | 27 | 100 | 27.00 |
| BioMistral-7B-DARE | Neurosurgery | 2024 | 33 | 99 | 33.33 |
| BioMistral-7B-DARE | Neurosurgery | 2025 | 25 | 100 | 25.00 |
| BioMistral-7B-DARE | Ophthalmology | 2018 | 11 | 97 | 11.34 |
| BioMistral-7B-DARE | Ophthalmology | 2019 | 26 | 98 | 26.53 |
| BioMistral-7B-DARE | Ophthalmology | 2020 | 35 | 98 | 35.71 |
| BioMistral-7B-DARE | Ophthalmology | 2022 | 24 | 97 | 24.74 |
| BioMistral-7B-DARE | Ophthalmology | 2023 | 29 | 99 | 29.29 |
| BioMistral-7B-DARE | Ophthalmology | 2024 | 21 | 100 | 21.00 |
| BioMistral-7B-DARE | Ophthalmology | 2025 | 34 | 100 | 34.00 |
| BioMistral-7B-DARE | Pediatrics | 2018 | 29 | 99 | 29.29 |
| BioMistral-7B-DARE | Pediatrics | 2019 | 32 | 100 | 32.00 |
| BioMistral-7B-DARE | Pediatrics | 2020 | 39 | 100 | 39.00 |
| BioMistral-7B-DARE | Pediatrics | 2022 | 33 | 99 | 33.33 |
| BioMistral-7B-DARE | Pediatrics | 2023 | 30 | 100 | 30.00 |
| BioMistral-7B-DARE | Pediatrics | 2024 | 27 | 100 | 27.00 |
| BioMistral-7B-DARE | Pediatrics | 2025 | 30 | 99 | 30.30 |
| BioMistral-7B-DARE | Test A | 2018 | 32 | 100 | 32.00 |
| BioMistral-7B-DARE | Test A | 2019 | 40 | 98 | 40.82 |
| BioMistral-7B-DARE | Test A | 2020 | 29 | 88 | 32.95 |
| BioMistral-7B-DARE | Test A | 2022 | 33 | 100 | 33.00 |
| BioMistral-7B-DARE | Test A | 2023 | 37 | 100 | 37.00 |

| BioMistral-7B-DARE | Test A | 2024 | 35 | 100 | 35.00 |
|---|---|---|---|---|---|
| BioMistral-7B-DARE | Test A | 2025 | 35 | 100 | 35.00 |
| BioMistral-7B-DARE | Test B | 2018 | 24 | 98 | 24.49 |
| BioMistral-7B-DARE | Test B | 2019 | 37 | 100 | 37.00 |
| BioMistral-7B-DARE | Test B | 2020 | 31 | 89 | 34.83 |
| BioMistral-7B-DARE | Test B | 2022 | 35 | 99 | 35.35 |
| BioMistral-7B-DARE | Test B | 2023 | 35 | 100 | 35.00 |
| BioMistral-7B-DARE | Test B | 2024 | 30 | 99 | 30.30 |
| BioMistral-7B-DARE | Test B | 2025 | 36 | 100 | 36.00 |
| BioMistral-7B-DARE | Psychiatry | 2018 | 42 | 99 | 42.42 |
| BioMistral-7B-DARE | Psychiatry | 2019 | 44 | 100 | 44.00 |
| BioMistral-7B-DARE | Psychiatry | 2020 | 51 | 100 | 51.00 |
| BioMistral-7B-DARE | Psychiatry | 2022 | 39 | 99 | 39.39 |
| BioMistral-7B-DARE | Psychiatry | 2023 | 46 | 100 | 46.00 |
| BioMistral-7B-DARE | Psychiatry | 2024 | 29 | 100 | 29.00 |
| BioMistral-7B-DARE | Psychiatry | 2025 | 38 | 100 | 38.00 |
| BioMistral-7B-DARE | Radiology | 2018 | 25 | 99 | 25.25 |
| BioMistral-7B-DARE | Radiology | 2019 | 40 | 99 | 40.40 |
| BioMistral-7B-DARE | Radiology | 2022 | 23 | 98 | 23.47 |
| BioMistral-7B-DARE | Radiology | 2023 | 32 | 100 | 32.00 |
| BioMistral-7B-DARE | Radiology | 2024 | 39 | 100 | 39.00 |
| BioMistral-7B-DARE | Radiology | 2025 | 30 | 97 | 30.93 |
| BioMistral-7B-DARE | Urology | 2018 | 33 | 100 | 33.00 |
| BioMistral-7B-DARE | Urology | 2020 | 41 | 100 | 41.00 |
| BioMistral-7B-DARE | Urology | 2022 | 22 | 98 | 22.45 |
| BioMistral-7B-DARE | Urology | 2023 | 31 | 100 | 31.00 |
| BioMistral-7B-DARE | Urology | 2024 | 26 | 96 | 27.08 |
| BioMistral-7B-DARE | Urology | 2025 | 32 | 100 | 32.00 |
| MediPhi-Instruct-3.8B | Pathological Anatomy and Pathology | 2018 | 36 | 100 | 36.00 |
| MediPhi-Instruct-3.8B | Pathological Anatomy and Pathology | 2019 | 29 | 100 | 29.00 |
| MediPhi-Instruct-3.8B | Pathological Anatomy and Pathology | 2020 | 33 | 99 | 33.33 |
| MediPhi-Instruct-3.8B | Pathological Anatomy and Pathology | 2023 | 45 | 100 | 45.00 |
| MediPhi-Instruct-3.8B | Pathological Anatomy and Pathology | 2024 | 39 | 100 | 39.00 |
| MediPhi-Instruct-3.8B | Pathological Anatomy and Pathology | 2025 | 41 | 100 | 41.00 |
| MediPhi-Instruct-3.8B | Anesthesiology | 2018 | 43 | 99 | 43.43 |
| MediPhi-Instruct-3.8B | Anesthesiology | 2019 | 48 | 100 | 48.00 |
| MediPhi-Instruct-3.8B | Anesthesiology | 2020 | 31 | 100 | 31.00 |
| MediPhi-Instruct-3.8B | Anesthesiology | 2022 | 40 | 100 | 40.00 |
| MediPhi-Instruct-3.8B | Anesthesiology | 2023 | 46 | 100 | 46.00 |
| MediPhi-Instruct-3.8B | Anesthesiology | 2024 | 44 | 99 | 44.44 |
| MediPhi-Instruct-3.8B | Anesthesiology | 2025 | 50 | 100 | 50.00 |
| MediPhi-Instruct-3.8B | General Surgery | 2018 | 30 | 100 | 30.00 |
| MediPhi-Instruct-3.8B | General Surgery | 2019 | 33 | 100 | 33.00 |
| MediPhi-Instruct-3.8B | General Surgery | 2020 | 34 | 100 | 34.00 |
| MediPhi-Instruct-3.8B | General Surgery | 2022 | 33 | 100 | 33.00 |
| MediPhi-Instruct-3.8B | General Surgery | 2023 | 41 | 100 | 41.00 |
| MediPhi-Instruct-3.8B | General Surgery | 2024 | 44 | 100 | 44.00 |
| MediPhi-Instruct-3.8B | General Surgery | 2025 | 35 | 100 | 35.00 |
| MediPhi-Instruct-3.8B | Thoracic and Cardiovascular Surgery | 2019 | 28 | 100 | 28.00 |
| MediPhi-Instruct-3.8B | Thoracic and Cardiovascular Surgery | 2022 | 41 | 100 | 41.00 |
| MediPhi-Instruct-3.8B | Thoracic and Cardiovascular Surgery | 2023 | 40 | 100 | 40.00 |
| MediPhi-Instruct-3.8B | Thoracic and Cardiovascular Surgery | 2024 | 47 | 100 | 47.00 |
| MediPhi-Instruct-3.8B | Thoracic and Cardiovascular Surgery | 2025 | 86 | 200 | 43.00 |
| MediPhi-Instruct-3.8B | Gynecology and Obstetrics | 2018 | 32 | 100 | 32.00 |
| MediPhi-Instruct-3.8B | Gynecology and Obstetrics | 2019 | 36 | 100 | 36.00 |
| MediPhi-Instruct-3.8B | Gynecology and Obstetrics | 2020 | 29 | 100 | 29.00 |
| MediPhi-Instruct-3.8B | Gynecology and Obstetrics | 2022 | 30 | 100 | 30.00 |
| MediPhi-Instruct-3.8B | Gynecology and Obstetrics | 2023 | 35 | 100 | 35.00 |
| MediPhi-Instruct-3.8B | Gynecology and Obstetrics | 2024 | 43 | 100 | 43.00 |
| MediPhi-Instruct-3.8B | Gynecology and Obstetrics | 2025 | 30 | 100 | 30.00 |
| MediPhi-Instruct-3.8B | Neurosurgery | 2022 | 28 | 100 | 28.00 |
| MediPhi-Instruct-3.8B | Neurosurgery | 2023 | 46 | 100 | 46.00 |
| MediPhi-Instruct-3.8B | Neurosurgery | 2024 | 41 | 100 | 41.00 |
| MediPhi-Instruct-3.8B | Neurosurgery | 2025 | 35 | 100 | 35.00 |
| MediPhi-Instruct-3.8B | Ophthalmology | 2018 | 42 | 100 | 42.00 |
| MediPhi-Instruct-3.8B | Ophthalmology | 2019 | 27 | 100 | 27.00 |
| MediPhi-Instruct-3.8B | Ophthalmology | 2020 | 24 | 100 | 24.00 |
| MediPhi-Instruct-3.8B | Ophthalmology | 2022 | 30 | 100 | 30.00 |
| MediPhi-Instruct-3.8B | Ophthalmology | 2023 | 46 | 100 | 46.00 |
| MediPhi-Instruct-3.8B | Ophthalmology | 2024 | 33 | 100 | 33.00 |
| MediPhi-Instruct-3.8B | Ophthalmology | 2025 | 37 | 100 | 37.00 |
| MediPhi-Instruct-3.8B | Pediatrics | 2018 | 45 | 100 | 45.00 |
| MediPhi-Instruct-3.8B | Pediatrics | 2019 | 39 | 100 | 39.00 |
| MediPhi-Instruct-3.8B | Pediatrics | 2020 | 35 | 100 | 35.00 |

| MediPhi-Instruct-3.8B | Pediatrics | 2022 | 39 | 100 | 39.00 |
|---|---|---|---|---|---|
| MediPhi-Instruct-3.8B | Pediatrics | 2023 | 45 | 100 | 45.00 |
| MediPhi-Instruct-3.8B | Pediatrics | 2024 | 45 | 100 | 45.00 |
| MediPhi-Instruct-3.8B | Pediatrics | 2025 | 34 | 100 | 34.00 |
| MediPhi-Instruct-3.8B | Test A | 2018 | 43 | 100 | 43.00 |
| MediPhi-Instruct-3.8B | Test A | 2019 | 38 | 100 | 38.00 |
| MediPhi-Instruct-3.8B | Test A | 2020 | 35 | 90 | 38.89 |
| MediPhi-Instruct-3.8B | Test A | 2022 | 42 | 100 | 42.00 |
| MediPhi-Instruct-3.8B | Test A | 2023 | 45 | 99 | 45.45 |
| MediPhi-Instruct-3.8B | Test A | 2024 | 46 | 100 | 46.00 |
| MediPhi-Instruct-3.8B | Test A | 2025 | 50 | 100 | 50.00 |
| MediPhi-Instruct-3.8B | Test B | 2018 | 39 | 100 | 39.00 |
| MediPhi-Instruct-3.8B | Test B | 2019 | 31 | 100 | 31.00 |
| MediPhi-Instruct-3.8B | Test B | 2020 | 37 | 90 | 41.11 |
| MediPhi-Instruct-3.8B | Test B | 2022 | 36 | 100 | 36.00 |
| MediPhi-Instruct-3.8B | Test B | 2023 | 40 | 100 | 40.00 |
| MediPhi-Instruct-3.8B | Test B | 2024 | 41 | 100 | 41.00 |
| MediPhi-Instruct-3.8B | Test B | 2025 | 42 | 100 | 42.00 |
| MediPhi-Instruct-3.8B | Psychiatry | 2018 | 55 | 100 | 55.00 |
| MediPhi-Instruct-3.8B | Psychiatry | 2019 | 45 | 100 | 45.00 |
| MediPhi-Instruct-3.8B | Psychiatry | 2020 | 45 | 100 | 45.00 |
| MediPhi-Instruct-3.8B | Psychiatry | 2022 | 41 | 100 | 41.00 |
| MediPhi-Instruct-3.8B | Psychiatry | 2023 | 58 | 100 | 58.00 |
| MediPhi-Instruct-3.8B | Psychiatry | 2024 | 52 | 100 | 52.00 |
| MediPhi-Instruct-3.8B | Psychiatry | 2025 | 52 | 100 | 52.00 |
| MediPhi-Instruct-3.8B | Radiology | 2018 | 42 | 100 | 42.00 |
| MediPhi-Instruct-3.8B | Radiology | 2019 | 41 | 100 | 41.00 |
| MediPhi-Instruct-3.8B | Radiology | 2022 | 26 | 100 | 26.00 |
| MediPhi-Instruct-3.8B | Radiology | 2023 | 41 | 100 | 41.00 |
| MediPhi-Instruct-3.8B | Radiology | 2024 | 39 | 100 | 39.00 |
| MediPhi-Instruct-3.8B | Radiology | 2025 | 32 | 100 | 32.00 |
| MediPhi-Instruct-3.8B | Urology | 2018 | 48 | 100 | 48.00 |
| MediPhi-Instruct-3.8B | Urology | 2020 | 28 | 100 | 28.00 |
| MediPhi-Instruct-3.8B | Urology | 2022 | 32 | 100 | 32.00 |
| MediPhi-Instruct-3.8B | Urology | 2023 | 50 | 100 | 50.00 |
| MediPhi-Instruct-3.8B | Urology | 2024 | 44 | 100 | 44.00 |
| MediPhi-Instruct-3.8B | Urology | 2025 | 38 | 100 | 38.00 |
| medgemma-4b-it | Pathological Anatomy and Pathology | 2018 | 46 | 99 | 46.46 |
| medgemma-4b-it | Pathological Anatomy and Pathology | 2019 | 46 | 100 | 46.00 |
| medgemma-4b-it | Pathological Anatomy and Pathology | 2020 | 41 | 100 | 41.00 |
| medgemma-4b-it | Pathological Anatomy and Pathology | 2023 | 41 | 99 | 41.41 |
| medgemma-4b-it | Pathological Anatomy and Pathology | 2024 | 46 | 100 | 46.00 |
| medgemma-4b-it | Pathological Anatomy and Pathology | 2025 | 47 | 100 | 47.00 |
| medgemma-4b-it | Anesthesiology | 2018 | 39 | 100 | 39.00 |
| medgemma-4b-it | Anesthesiology | 2019 | 52 | 100 | 52.00 |
| medgemma-4b-it | Anesthesiology | 2020 | 44 | 100 | 44.00 |
| medgemma-4b-it | Anesthesiology | 2022 | 40 | 100 | 40.00 |
| medgemma-4b-it | Anesthesiology | 2023 | 47 | 100 | 47.00 |
| medgemma-4b-it | Anesthesiology | 2024 | 44 | 100 | 44.00 |
| medgemma-4b-it | Anesthesiology | 2025 | 43 | 100 | 43.00 |
| medgemma-4b-it | General Surgery | 2018 | 42 | 100 | 42.00 |
| medgemma-4b-it | General Surgery | 2019 | 46 | 100 | 46.00 |
| medgemma-4b-it | General Surgery | 2020 | 43 | 100 | 43.00 |
| medgemma-4b-it | General Surgery | 2022 | 43 | 100 | 43.00 |
| medgemma-4b-it | General Surgery | 2023 | 41 | 100 | 41.00 |
| medgemma-4b-it | General Surgery | 2024 | 39 | 100 | 39.00 |
| medgemma-4b-it | General Surgery | 2025 | 46 | 100 | 46.00 |
| medgemma-4b-it | Thoracic and Cardiovascular Surgery | 2019 | 31 | 99 | 31.31 |
| medgemma-4b-it | Thoracic and Cardiovascular Surgery | 2022 | 45 | 100 | 45.00 |
| medgemma-4b-it | Thoracic and Cardiovascular Surgery | 2023 | 46 | 99 | 46.46 |
| medgemma-4b-it | Thoracic and Cardiovascular Surgery | 2024 | 47 | 100 | 47.00 |
| medgemma-4b-it | Thoracic and Cardiovascular Surgery | 2025 | 88 | 200 | 44.00 |
| medgemma-4b-it | Gynecology and Obstetrics | 2018 | 36 | 100 | 36.00 |
| medgemma-4b-it | Gynecology and Obstetrics | 2019 | 41 | 100 | 41.00 |
| medgemma-4b-it | Gynecology and Obstetrics | 2020 | 36 | 100 | 36.00 |
| medgemma-4b-it | Gynecology and Obstetrics | 2022 | 34 | 100 | 34.00 |
| medgemma-4b-it | Gynecology and Obstetrics | 2023 | 37 | 100 | 37.00 |
| medgemma-4b-it | Gynecology and Obstetrics | 2024 | 42 | 100 | 42.00 |
| medgemma-4b-it | Gynecology and Obstetrics | 2025 | 34 | 100 | 34.00 |
| medgemma-4b-it | Neurosurgery | 2022 | 35 | 100 | 35.00 |
| medgemma-4b-it | Neurosurgery | 2023 | 45 | 100 | 45.00 |
| medgemma-4b-it | Neurosurgery | 2024 | 52 | 100 | 52.00 |
| medgemma-4b-it | Neurosurgery | 2025 | 42 | 100 | 42.00 |
| medgemma-4b-it | Ophthalmology | 2018 | 35 | 100 | 35.00 |
| medgemma-4b-it | Ophthalmology | 2019 | 38 | 100 | 38.00 |
| medgemma-4b-it | Ophthalmology | 2020 | 37 | 100 | 37.00 |
| medgemma-4b-it | Ophthalmology | 2022 | 37 | 100 | 37.00 |
| medgemma-4b-it | Ophthalmology | 2023 | 34 | 99 | 34.34 |
| medgemma-4b-it | Ophthalmology | 2024 | 35 | 100 | 35.00 |
| medgemma-4b-it | Ophthalmology | 2025 | 38 | 98 | 38.78 |
| medgemma-4b-it | Pediatrics | 2018 | 43 | 99 | 43.43 |
| medgemma-4b-it | Pediatrics | 2019 | 54 | 100 | 54.00 |
| medgemma-4b-it | Pediatrics | 2020 | 46 | 100 | 46.00 |
| medgemma-4b-it | Pediatrics | 2022 | 46 | 100 | 46.00 |
| medgemma-4b-it | Pediatrics | 2023 | 52 | 100 | 52.00 |
| medgemma-4b-it | Pediatrics | 2024 | 46 | 99 | 46.46 |
| medgemma-4b-it | Pediatrics | 2025 | 45 | 100 | 45.00 |
| medgemma-4b-it | Test A | 2018 | 52 | 100 | 52.00 |
| medgemma-4b-it | Test A | 2019 | 58 | 99 | 58.59 |
| medgemma-4b-it | Test A | 2020 | 46 | 90 | 51.11 |
| medgemma-4b-it | Test A | 2022 | 56 | 100 | 56.00 |
| medgemma-4b-it | Test A | 2023 | 48 | 100 | 48.00 |
| medgemma-4b-it | Test A | 2024 | 57 | 100 | 57.00 |
| medgemma-4b-it | Test A | 2025 | 45 | 100 | 45.00 |
| medgemma-4b-it | Test B | 2018 | 54 | 100 | 54.00 |
| medgemma-4b-it | Test B | 2019 | 48 | 100 | 48.00 |
| medgemma-4b-it | Test B | 2020 | 47 | 90 | 52.22 |
| medgemma-4b-it | Test B | 2022 | 52 | 100 | 52.00 |
| medgemma-4b-it | Test B | 2023 | 38 | 100 | 38.00 |
| medgemma-4b-it | Test B | 2024 | 43 | 100 | 43.00 |
| medgemma-4b-it | Test B | 2025 | 55 | 100 | 55.00 |
| medgemma-4b-it | Psychiatry | 2018 | 52 | 99 | 52.53 |
| medgemma-4b-it | Psychiatry | 2019 | 56 | 100 | 56.00 |
| medgemma-4b-it | Psychiatry | 2020 | 66 | 100 | 66.00 |
| medgemma-4b-it | Psychiatry | 2022 | 60 | 100 | 60.00 |
| medgemma-4b-it | Psychiatry | 2023 | 61 | 100 | 61.00 |
| medgemma-4b-it | Psychiatry | 2024 | 58 | 100 | 58.00 |
| medgemma-4b-it | Psychiatry | 2025 | 65 | 100 | 65.00 |
| medgemma-4b-it | Radiology | 2018 | 40 | 100 | 40.00 |
| medgemma-4b-it | Radiology | 2019 | 47 | 100 | 47.00 |
| medgemma-4b-it | Radiology | 2022 | 36 | 100 | 36.00 |
| medgemma-4b-it | Radiology | 2023 | 43 | 100 | 43.00 |
| medgemma-4b-it | Radiology | 2024 | 43 | 100 | 43.00 |

| medgemma-4b-it | Radiology | 2025 | 39 | 100 | 39.00 |
|---|---|---|---|---|---|
| medgemma-4b-it | Urology | 2018 | 57 | 100 | 57.00 |
| medgemma-4b-it | Urology | 2020 | 42 | 100 | 42.00 |
| medgemma-4b-it | Urology | 2022 | 41 | 100 | 41.00 |
| medgemma-4b-it | Urology | 2023 | 50 | 99 | 50.51 |
| medgemma-4b-it | Urology | 2024 | 37 | 100 | 37.00 |
| medgemma-4b-it | Urology | 2025 | 34 | 100 | 34.00 |
| medgemma-1.5-4b-it | Pathological Anatomy and Pathology | 2018 | 63 | 96 | 65.62 |
| medgemma-1.5-4b-it | Pathological Anatomy and Pathology | 2019 | 51 | 96 | 53.12 |
| medgemma-1.5-4b-it | Pathological Anatomy and Pathology | 2020 | 54 | 99 | 54.55 |
| medgemma-1.5-4b-it | Pathological Anatomy and Pathology | 2023 | 64 | 99 | 64.65 |
| medgemma-1.5-4b-it | Pathological Anatomy and Pathology | 2024 | 61 | 99 | 61.62 |
| medgemma-1.5-4b-it | Pathological Anatomy and Pathology | 2025 | 51 | 98 | 52.04 |
| medgemma-1.5-4b-it | Anesthesiology | 2018 | 55 | 96 | 57.29 |
| medgemma-1.5-4b-it | Anesthesiology | 2019 | 65 | 97 | 67.01 |
| medgemma-1.5-4b-it | Anesthesiology | 2020 | 50 | 94 | 53.19 |
| medgemma-1.5-4b-it | Anesthesiology | 2022 | 48 | 96 | 50.00 |
| medgemma-1.5-4b-it | Anesthesiology | 2023 | 54 | 96 | 56.25 |
| medgemma-1.5-4b-it | Anesthesiology | 2024 | 60 | 99 | 60.61 |
| medgemma-1.5-4b-it | Anesthesiology | 2025 | 57 | 96 | 59.38 |
| medgemma-1.5-4b-it | General Surgery | 2018 | 43 | 97 | 44.33 |
| medgemma-1.5-4b-it | General Surgery | 2019 | 57 | 99 | 57.58 |
| medgemma-1.5-4b-it | General Surgery | 2020 | 50 | 94 | 53.19 |
| medgemma-1.5-4b-it | General Surgery | 2022 | 51 | 95 | 53.68 |
| medgemma-1.5-4b-it | General Surgery | 2023 | 52 | 99 | 52.53 |
| medgemma-1.5-4b-it | General Surgery | 2024 | 50 | 100 | 50.00 |
| medgemma-1.5-4b-it | General Surgery | 2025 | 56 | 100 | 56.00 |
| medgemma-1.5-4b-it | Thoracic and Cardiovascular Surgery | 2019 | 38 | 99 | 38.38 |
| medgemma-1.5-4b-it | Thoracic and Cardiovascular Surgery | 2022 | 58 | 96 | 60.42 |
| medgemma-1.5-4b-it | Thoracic and Cardiovascular Surgery | 2023 | 51 | 95 | 53.68 |
| medgemma-1.5-4b-it | Thoracic and Cardiovascular Surgery | 2024 | 47 | 96 | 48.96 |
| medgemma-1.5-4b-it | Thoracic and Cardiovascular Surgery | 2025 | 112 | 194 | 57.73 |
| medgemma-1.5-4b-it | Gynecology and Obstetrics | 2018 | 55 | 97 | 56.70 |
| medgemma-1.5-4b-it | Gynecology and Obstetrics | 2019 | 53 | 97 | 54.64 |
| medgemma-1.5-4b-it | Gynecology and Obstetrics | 2020 | 46 | 96 | 47.92 |
| medgemma-1.5-4b-it | Gynecology and Obstetrics | 2022 | 46 | 99 | 46.46 |
| medgemma-1.5-4b-it | Gynecology and Obstetrics | 2023 | 48 | 99 | 48.48 |
| medgemma-1.5-4b-it | Gynecology and Obstetrics | 2024 | 47 | 99 | 47.47 |
| medgemma-1.5-4b-it | Gynecology and Obstetrics | 2025 | 51 | 95 | 53.68 |
| medgemma-1.5-4b-it | Neurosurgery | 2022 | 50 | 97 | 51.55 |
| medgemma-1.5-4b-it | Neurosurgery | 2023 | 50 | 95 | 52.63 |
| medgemma-1.5-4b-it | Neurosurgery | 2024 | 49 | 98 | 50.00 |
| medgemma-1.5-4b-it | Neurosurgery | 2025 | 53 | 96 | 55.21 |
| medgemma-1.5-4b-it | Ophthalmology | 2018 | 47 | 98 | 47.96 |
| medgemma-1.5-4b-it | Ophthalmology | 2019 | 41 | 97 | 42.27 |
| medgemma-1.5-4b-it | Ophthalmology | 2020 | 47 | 96 | 48.96 |
| medgemma-1.5-4b-it | Ophthalmology | 2022 | 39 | 99 | 39.39 |
| medgemma-1.5-4b-it | Ophthalmology | 2023 | 47 | 96 | 48.96 |
| medgemma-1.5-4b-it | Ophthalmology | 2024 | 42 | 100 | 42.00 |
| medgemma-1.5-4b-it | Ophthalmology | 2025 | 51 | 98 | 52.04 |
| medgemma-1.5-4b-it | Pediatrics | 2018 | 54 | 97 | 55.67 |
| medgemma-1.5-4b-it | Pediatrics | 2019 | 63 | 93 | 67.74 |
| medgemma-1.5-4b-it | Pediatrics | 2020 | 53 | 96 | 55.21 |
| medgemma-1.5-4b-it | Pediatrics | 2022 | 57 | 97 | 58.76 |
| medgemma-1.5-4b-it | Pediatrics | 2023 | 57 | 97 | 58.76 |
| medgemma-1.5-4b-it | Pediatrics | 2024 | 63 | 96 | 65.62 |
| medgemma-1.5-4b-it | Pediatrics | 2025 | 59 | 96 | 61.46 |
| medgemma-1.5-4b-it | Test A | 2018 | 66 | 97 | 68.04 |
| medgemma-1.5-4b-it | Test A | 2019 | 66 | 99 | 66.67 |
| medgemma-1.5-4b-it | Test A | 2020 | 51 | 88 | 57.95 |
| medgemma-1.5-4b-it | Test A | 2022 | 68 | 95 | 71.58 |
| medgemma-1.5-4b-it | Test A | 2023 | 62 | 98 | 63.27 |
| medgemma-1.5-4b-it | Test A | 2024 | 62 | 98 | 63.27 |
| medgemma-1.5-4b-it | Test A | 2025 | 57 | 98 | 58.16 |
| medgemma-1.5-4b-it | Test B | 2018 | 63 | 97 | 64.95 |
| medgemma-1.5-4b-it | Test B | 2019 | 54 | 93 | 58.06 |
| medgemma-1.5-4b-it | Test B | 2020 | 57 | 88 | 64.77 |
| medgemma-1.5-4b-it | Test B | 2022 | 58 | 96 | 60.42 |
| medgemma-1.5-4b-it | Test B | 2023 | 54 | 98 | 55.10 |
| medgemma-1.5-4b-it | Test B | 2024 | 51 | 96 | 53.12 |
| medgemma-1.5-4b-it | Test B | 2025 | 59 | 98 | 60.20 |
| medgemma-1.5-4b-it | Psychiatry | 2018 | 65 | 96 | 67.71 |
| medgemma-1.5-4b-it | Psychiatry | 2019 | 63 | 98 | 64.29 |
| medgemma-1.5-4b-it | Psychiatry | 2020 | 73 | 99 | 73.74 |
| medgemma-1.5-4b-it | Psychiatry | 2022 | 65 | 98 | 66.33 |

| medgemma-1.5-4b-it | Psychiatry | 2023 | 68 | 95 | 71.58 |
|---|---|---|---|---|---|
| medgemma-1.5-4b-it | Psychiatry | 2024 | 69 | 100 | 69.00 |
| medgemma-1.5-4b-it | Psychiatry | 2025 | 70 | 97 | 72.16 |
| medgemma-1.5-4b-it | Radiology | 2018 | 56 | 98 | 57.14 |
| medgemma-1.5-4b-it | Radiology | 2019 | 58 | 99 | 58.59 |
| medgemma-1.5-4b-it | Radiology | 2022 | 44 | 97 | 45.36 |
| medgemma-1.5-4b-it | Radiology | 2023 | 57 | 98 | 58.16 |
| medgemma-1.5-4b-it | Radiology | 2024 | 50 | 96 | 52.08 |
| medgemma-1.5-4b-it | Radiology | 2025 | 49 | 96 | 51.04 |
| medgemma-1.5-4b-it | Urology | 2018 | 52 | 99 | 52.53 |
| medgemma-1.5-4b-it | Urology | 2020 | 57 | 100 | 57.00 |
| medgemma-1.5-4b-it | Urology | 2022 | 53 | 100 | 53.00 |
| medgemma-1.5-4b-it | Urology | 2023 | 59 | 98 | 60.20 |
| medgemma-1.5-4b-it | Urology | 2024 | 52 | 99 | 52.53 |
| medgemma-1.5-4b-it | Urology | 2025 | 55 | 98 | 56.12 |
| JSL-MedLlama-3-8B-v2.0 | Pathological Anatomy and Pathology | 2018 | 46 | 92 | 50.00 |
| JSL-MedLlama-3-8B-v2.0 | Pathological Anatomy and Pathology | 2019 | 36 | 82 | 43.90 |
| JSL-MedLlama-3-8B-v2.0 | Pathological Anatomy and Pathology | 2020 | 43 | 93 | 46.24 |
| JSL-MedLlama-3-8B-v2.0 | Pathological Anatomy and Pathology | 2023 | 51 | 90 | 56.67 |
| JSL-MedLlama-3-8B-v2.0 | Pathological Anatomy and Pathology | 2024 | 54 | 92 | 58.70 |
| JSL-MedLlama-3-8B-v2.0 | Pathological Anatomy and Pathology | 2025 | 50 | 99 | 50.51 |
| JSL-MedLlama-3-8B-v2.0 | Anesthesiology | 2018 | 47 | 89 | 52.81 |
| JSL-MedLlama-3-8B-v2.0 | Anesthesiology | 2019 | 39 | 82 | 47.56 |
| JSL-MedLlama-3-8B-v2.0 | Anesthesiology | 2020 | 32 | 79 | 40.51 |
| JSL-MedLlama-3-8B-v2.0 | Anesthesiology | 2022 | 41 | 86 | 47.67 |
| JSL-MedLlama-3-8B-v2.0 | Anesthesiology | 2023 | 45 | 93 | 48.39 |
| JSL-MedLlama-3-8B-v2.0 | Anesthesiology | 2024 | 47 | 89 | 52.81 |
| JSL-MedLlama-3-8B-v2.0 | Anesthesiology | 2025 | 49 | 93 | 52.69 |
| JSL-MedLlama-3-8B-v2.0 | General Surgery | 2018 | 37 | 90 | 41.11 |
| JSL-MedLlama-3-8B-v2.0 | General Surgery | 2019 | 33 | 89 | 37.08 |
| JSL-MedLlama-3-8B-v2.0 | General Surgery | 2020 | 36 | 93 | 38.71 |
| JSL-MedLlama-3-8B-v2.0 | General Surgery | 2022 | 37 | 92 | 40.22 |
| JSL-MedLlama-3-8B-v2.0 | General Surgery | 2023 | 47 | 93 | 50.54 |
| JSL-MedLlama-3-8B-v2.0 | General Surgery | 2024 | 47 | 93 | 50.54 |
| JSL-MedLlama-3-8B-v2.0 | General Surgery | 2025 | 49 | 91 | 53.85 |
| JSL-MedLlama-3-8B-v2.0 | Thoracic and Cardiovascular Surgery | 2019 | 23 | 88 | 26.14 |

| JSL-MedLlama-3-8B-v2.0 | Thoracic and Cardiovascular Surgery | 2022 | 35 | 80 | 43.75 |
|---|---|---|---|---|---|
| JSL-MedLlama-3-8B-v2.0 | Thoracic and Cardiovascular Surgery | 2023 | 47 | 84 | 55.95 |
| JSL-MedLlama-3-8B-v2.0 | Thoracic and Cardiovascular Surgery | 2024 | 43 | 91 | 47.25 |
| JSL-MedLlama-3-8B-v2.0 | Thoracic and Cardiovascular Surgery | 2025 | 86 | 184 | 46.74 |
| JSL-MedLlama-3-8B-v2.0 | Gynecology and Obstetrics | 2018 | 39 | 92 | 42.39 |
| JSL-MedLlama-3-8B-v2.0 | Gynecology and Obstetrics | 2019 | 38 | 88 | 43.18 |
| JSL-MedLlama-3-8B-v2.0 | Gynecology and Obstetrics | 2020 | 28 | 91 | 30.77 |
| JSL-MedLlama-3-8B-v2.0 | Gynecology and Obstetrics | 2022 | 36 | 92 | 39.13 |
| JSL-MedLlama-3-8B-v2.0 | Gynecology and Obstetrics | 2023 | 36 | 95 | 37.89 |
| JSL-MedLlama-3-8B-v2.0 | Gynecology and Obstetrics | 2024 | 44 | 89 | 49.44 |
| JSL-MedLlama-3-8B-v2.0 | Gynecology and Obstetrics | 2025 | 50 | 95 | 52.63 |
| JSL-MedLlama-3-8B-v2.0 | Neurosurgery | 2022 | 40 | 95 | 42.11 |
| JSL-MedLlama-3-8B-v2.0 | Neurosurgery | 2023 | 49 | 93 | 52.69 |
| JSL-MedLlama-3-8B-v2.0 | Neurosurgery | 2024 | 46 | 90 | 51.11 |
| JSL-MedLlama-3-8B-v2.0 | Neurosurgery | 2025 | 45 | 95 | 47.37 |
| JSL-MedLlama-3-8B-v2.0 | Ophthalmology | 2018 | 36 | 93 | 38.71 |
| JSL-MedLlama-3-8B-v2.0 | Ophthalmology | 2019 | 35 | 91 | 38.46 |
| JSL-MedLlama-3-8B-v2.0 | Ophthalmology | 2020 | 31 | 84 | 36.90 |
| JSL-MedLlama-3-8B-v2.0 | Ophthalmology | 2022 | 33 | 89 | 37.08 |
| JSL-MedLlama-3-8B-v2.0 | Ophthalmology | 2023 | 42 | 94 | 44.68 |
| JSL-MedLlama-3-8B-v2.0 | Ophthalmology | 2024 | 31 | 90 | 34.44 |
| JSL-MedLlama-3-8B-v2.0 | Ophthalmology | 2025 | 34 | 95 | 35.79 |
| JSL-MedLlama-3-8B-v2.0 | Pediatrics | 2018 | 43 | 91 | 47.25 |
| JSL-MedLlama-3-8B-v2.0 | Pediatrics | 2019 | 44 | 90 | 48.89 |
| JSL-MedLlama-3-8B-v2.0 | Pediatrics | 2020 | 40 | 89 | 44.94 |
| JSL-MedLlama-3-8B-v2.0 | Pediatrics | 2022 | 41 | 89 | 46.07 |
| JSL-MedLlama-3-8B-v2.0 | Pediatrics | 2023 | 50 | 95 | 52.63 |
| JSL-MedLlama-3-8B-v2.0 | Pediatrics | 2024 | 47 | 93 | 50.54 |
| JSL-MedLlama-3-8B-v2.0 | Pediatrics | 2025 | 50 | 93 | 53.76 |
| JSL-MedLlama-3-8B-v2.0 | Test A | 2018 | 53 | 92 | 57.61 |
| JSL-MedLlama-3-8B-v2.0 | Test A | 2019 | 50 | 88 | 56.82 |
| JSL-MedLlama-3-8B-v2.0 | Test A | 2020 | 40 | 82 | 48.78 |
| JSL-MedLlama-3-8B-v2.0 | Test A | 2022 | 49 | 92 | 53.26 |
| JSL-MedLlama-3-8B-v2.0 | Test A | 2023 | 52 | 94 | 55.32 |
| JSL-MedLlama-3-8B-v2.0 | Test A | 2024 | 56 | 91 | 61.54 |
| JSL-MedLlama-3-8B-v2.0 | Test A | 2025 | 54 | 94 | 57.45 |
| JSL-MedLlama-3-8B-v2.0 | Test B | 2018 | 55 | 91 | 60.44 |
| JSL-MedLlama-3-8B-v2.0 | Test B | 2019 | 46 | 86 | 53.49 |

| JSL-MedLlama-3-8B-v2.0 | Test B | 2020 | 41 | 80 | 51.25 |
|---|---|---|---|---|---|
| JSL-MedLlama-3-8B-v2.0 | Test B | 2022 | 39 | 94 | 41.49 |
| JSL-MedLlama-3-8B-v2.0 | Test B | 2023 | 49 | 93 | 52.69 |
| JSL-MedLlama-3-8B-v2.0 | Test B | 2024 | 42 | 95 | 44.21 |
| JSL-MedLlama-3-8B-v2.0 | Test B | 2025 | 54 | 90 | 60.00 |
| JSL-MedLlama-3-8B-v2.0 | Psychiatry | 2018 | 60 | 93 | 64.52 |
| JSL-MedLlama-3-8B-v2.0 | Psychiatry | 2019 | 57 | 87 | 65.52 |
| JSL-MedLlama-3-8B-v2.0 | Psychiatry | 2020 | 67 | 94 | 71.28 |
| JSL-MedLlama-3-8B-v2.0 | Psychiatry | 2022 | 44 | 87 | 50.57 |
| JSL-MedLlama-3-8B-v2.0 | Psychiatry | 2023 | 68 | 91 | 74.73 |
| JSL-MedLlama-3-8B-v2.0 | Psychiatry | 2024 | 68 | 92 | 73.91 |
| JSL-MedLlama-3-8B-v2.0 | Psychiatry | 2025 | 66 | 95 | 69.47 |
| JSL-MedLlama-3-8B-v2.0 | Radiology | 2018 | 44 | 87 | 50.57 |
| JSL-MedLlama-3-8B-v2.0 | Radiology | 2019 | 38 | 90 | 42.22 |
| JSL-MedLlama-3-8B-v2.0 | Radiology | 2022 | 32 | 95 | 33.68 |
| JSL-MedLlama-3-8B-v2.0 | Radiology | 2023 | 38 | 94 | 40.43 |
| JSL-MedLlama-3-8B-v2.0 | Radiology | 2024 | 48 | 94 | 51.06 |
| JSL-MedLlama-3-8B-v2.0 | Radiology | 2025 | 40 | 94 | 42.55 |
| JSL-MedLlama-3-8B-v2.0 | Urology | 2018 | 55 | 90 | 61.11 |
| JSL-MedLlama-3-8B-v2.0 | Urology | 2020 | 34 | 92 | 36.96 |
| JSL-MedLlama-3-8B-v2.0 | Urology | 2022 | 30 | 92 | 32.61 |
| JSL-MedLlama-3-8B-v2.0 | Urology | 2023 | 52 | 94 | 55.32 |
| JSL-MedLlama-3-8B-v2.0 | Urology | 2024 | 43 | 92 | 46.74 |
| JSL-MedLlama-3-8B-v2.0 | Urology | 2025 | 40 | 92 | 43.48 |
| Llama3-OpenBioLLM-70B | Pathological Anatomy and Pathology | 2018 | 74 | 100 | 74.00 |
| Llama3-OpenBioLLM-70B | Pathological Anatomy and Pathology | 2019 | 64 | 100 | 64.00 |
| Llama3-OpenBioLLM-70B | Pathological Anatomy and Pathology | 2020 | 70 | 100 | 70.00 |
| Llama3-OpenBioLLM-70B | Pathological Anatomy and Pathology | 2023 | 77 | 100 | 77.00 |
| Llama3-OpenBioLLM-70B | Pathological Anatomy and Pathology | 2024 | 77 | 100 | 77.00 |
| Llama3-OpenBioLLM-70B | Pathological Anatomy and Pathology | 2025 | 70 | 100 | 70.00 |
| Llama3-OpenBioLLM-70B | Anesthesiology | 2018 | 69 | 100 | 69.00 |
| Llama3-OpenBioLLM-70B | Anesthesiology | 2019 | 70 | 100 | 70.00 |
| Llama3-OpenBioLLM-70B | Anesthesiology | 2020 | 64 | 100 | 64.00 |
| Llama3-OpenBioLLM-70B | Anesthesiology | 2022 | 66 | 100 | 66.00 |
| Llama3-OpenBioLLM-70B | Anesthesiology | 2023 | 65 | 100 | 65.00 |
| Llama3-OpenBioLLM-70B | Anesthesiology | 2024 | 75 | 100 | 75.00 |
| Llama3-OpenBioLLM-70B | Anesthesiology | 2025 | 74 | 100 | 74.00 |
| Llama3-OpenBioLLM-70B | General Surgery | 2018 | 62 | 100 | 62.00 |
| Llama3-OpenBioLLM-70B | General Surgery | 2019 | 60 | 100 | 60.00 |
| Llama3-OpenBioLLM-70B | General Surgery | 2020 | 63 | 100 | 63.00 |
| Llama3-OpenBioLLM-70B | General Surgery | 2022 | 64 | 100 | 64.00 |
| Llama3-OpenBioLLM-70B | General Surgery | 2023 | 74 | 100 | 74.00 |
| Llama3-OpenBioLLM-70B | General Surgery | 2024 | 73 | 100 | 73.00 |
| Llama3-OpenBioLLM-70B | General Surgery | 2025 | 71 | 100 | 71.00 |
| Llama3-OpenBioLLM-70B | Thoracic and Cardiovascular Surgery | 2019 | 51 | 100 | 51.00 |
| Llama3-OpenBioLLM-70B | Thoracic and Cardiovascular Surgery | 2022 | 69 | 100 | 69.00 |
| Llama3-OpenBioLLM-70B | Thoracic and Cardiovascular Surgery | 2023 | 71 | 100 | 71.00 |
| Llama3-OpenBioLLM-70B | Thoracic and Cardiovascular Surgery | 2024 | 69 | 100 | 69.00 |
| Llama3-OpenBioLLM-70B | Thoracic and Cardiovascular Surgery | 2025 | 146 | 200 | 73.00 |
| Llama3-OpenBioLLM-70B | Gynecology and Obstetrics | 2018 | 69 | 100 | 69.00 |
| Llama3-OpenBioLLM-70B | Gynecology and Obstetrics | 2019 | 68 | 100 | 68.00 |
| Llama3-OpenBioLLM-70B | Gynecology and Obstetrics | 2020 | 63 | 100 | 63.00 |
| Llama3-OpenBioLLM-70B | Gynecology and Obstetrics | 2022 | 62 | 100 | 62.00 |
| Llama3-OpenBioLLM-70B | Gynecology and Obstetrics | 2023 | 63 | 100 | 63.00 |
| Llama3-OpenBioLLM-70B | Gynecology and Obstetrics | 2024 | 64 | 100 | 64.00 |
| Llama3-OpenBioLLM-70B | Gynecology and Obstetrics | 2025 | 68 | 100 | 68.00 |
| Llama3-OpenBioLLM-70B | Neurosurgery | 2022 | 62 | 100 | 62.00 |
| Llama3-OpenBioLLM-70B | Neurosurgery | 2023 | 74 | 100 | 74.00 |
| Llama3-OpenBioLLM-70B | Neurosurgery | 2024 | 75 | 100 | 75.00 |
| Llama3-OpenBioLLM-70B | Neurosurgery | 2025 | 57 | 100 | 57.00 |
| Llama3-OpenBioLLM-70B | Ophthalmology | 2018 | 57 | 100 | 57.00 |
| Llama3-OpenBioLLM-70B | Ophthalmology | 2019 | 63 | 100 | 63.00 |
| Llama3-OpenBioLLM-70B | Ophthalmology | 2020 | 70 | 100 | 70.00 |
| Llama3-OpenBioLLM-70B | Ophthalmology | 2022 | 70 | 100 | 70.00 |
| Llama3-OpenBioLLM-70B | Ophthalmology | 2023 | 66 | 100 | 66.00 |

| Llama3-OpenBioLLM-70B | Ophthalmology | 2024 | 64 | 100 | 64.00 |
|---|---|---|---|---|---|
| Llama3-OpenBioLLM-70B | Ophthalmology | 2025 | 65 | 100 | 65.00 |
| Llama3-OpenBioLLM-70B | Pediatrics | 2018 | 63 | 100 | 63.00 |
| Llama3-OpenBioLLM-70B | Pediatrics | 2019 | 70 | 100 | 70.00 |
| Llama3-OpenBioLLM-70B | Pediatrics | 2020 | 58 | 100 | 58.00 |
| Llama3-OpenBioLLM-70B | Pediatrics | 2022 | 71 | 100 | 71.00 |
| Llama3-OpenBioLLM-70B | Pediatrics | 2023 | 72 | 100 | 72.00 |
| Llama3-OpenBioLLM-70B | Pediatrics | 2024 | 70 | 100 | 70.00 |
| Llama3-OpenBioLLM-70B | Pediatrics | 2025 | 74 | 100 | 74.00 |
| Llama3-OpenBioLLM-70B | Test A | 2018 | 75 | 100 | 75.00 |
| Llama3-OpenBioLLM-70B | Test A | 2019 | 69 | 100 | 69.00 |
| Llama3-OpenBioLLM-70B | Test A | 2020 | 68 | 90 | 75.56 |
| Llama3-OpenBioLLM-70B | Test A | 2022 | 78 | 100 | 78.00 |
| Llama3-OpenBioLLM-70B | Test A | 2023 | 80 | 100 | 80.00 |
| Llama3-OpenBioLLM-70B | Test A | 2024 | 78 | 100 | 78.00 |
| Llama3-OpenBioLLM-70B | Test A | 2025 | 79 | 100 | 79.00 |
| Llama3-OpenBioLLM-70B | Test B | 2018 | 73 | 100 | 73.00 |
| Llama3-OpenBioLLM-70B | Test B | 2019 | 72 | 100 | 72.00 |
| Llama3-OpenBioLLM-70B | Test B | 2020 | 74 | 90 | 82.22 |
| Llama3-OpenBioLLM-70B | Test B | 2022 | 76 | 100 | 76.00 |
| Llama3-OpenBioLLM-70B | Test B | 2023 | 69 | 100 | 69.00 |
| Llama3-OpenBioLLM-70B | Test B | 2024 | 78 | 100 | 78.00 |
| Llama3-OpenBioLLM-70B | Test B | 2025 | 77 | 100 | 77.00 |
| Llama3-OpenBioLLM-70B | Psychiatry | 2018 | 79 | 100 | 79.00 |
| Llama3-OpenBioLLM-70B | Psychiatry | 2019 | 87 | 100 | 87.00 |
| Llama3-OpenBioLLM-70B | Psychiatry | 2020 | 91 | 100 | 91.00 |
| Llama3-OpenBioLLM-70B | Psychiatry | 2022 | 78 | 100 | 78.00 |
| Llama3-OpenBioLLM-70B | Psychiatry | 2023 | 83 | 100 | 83.00 |
| Llama3-OpenBioLLM-70B | Psychiatry | 2024 | 92 | 100 | 92.00 |
| Llama3-OpenBioLLM-70B | Psychiatry | 2025 | 89 | 100 | 89.00 |

| Llama3-OpenBioLLM-70B | Radiology | 2018 | 71 | 100 | 71.00 |
|---|---|---|---|---|---|
| Llama3-OpenBioLLM-70B | Radiology | 2019 | 73 | 100 | 73.00 |
| Llama3-OpenBioLLM-70B | Radiology | 2022 | 67 | 100 | 67.00 |
| Llama3-OpenBioLLM-70B | Radiology | 2023 | 71 | 100 | 71.00 |
| Llama3-OpenBioLLM-70B | Radiology | 2024 | 72 | 100 | 72.00 |
| Llama3-OpenBioLLM-70B | Radiology | 2025 | 62 | 100 | 62.00 |
| Llama3-OpenBioLLM-70B | Urology | 2018 | 83 | 100 | 83.00 |
| Llama3-OpenBioLLM-70B | Urology | 2020 | 63 | 100 | 63.00 |
| Llama3-OpenBioLLM-70B | Urology | 2022 | 68 | 100 | 68.00 |
| Llama3-OpenBioLLM-70B | Urology | 2023 | 75 | 100 | 75.00 |
| Llama3-OpenBioLLM-70B | Urology | 2024 | 67 | 100 | 67.00 |
| Llama3-OpenBioLLM-70B | Urology | 2025 | 69 | 100 | 69.00 |
| medgemma-27b-text-it | Pathological Anatomy and Pathology | 2018 | 82 | 100 | 82.00 |
| medgemma-27b-text-it | Pathological Anatomy and Pathology | 2019 | 77 | 100 | 77.00 |
| medgemma-27b-text-it | Pathological Anatomy and Pathology | 2020 | 88 | 100 | 88.00 |
| medgemma-27b-text-it | Pathological Anatomy and Pathology | 2023 | 86 | 100 | 86.00 |
| medgemma-27b-text-it | Pathological Anatomy and Pathology | 2024 | 92 | 100 | 92.00 |
| medgemma-27b-text-it | Pathological Anatomy and Pathology | 2025 | 85 | 100 | 85.00 |
| medgemma-27b-text-it | Anesthesiology | 2018 | 85 | 100 | 85.00 |
| medgemma-27b-text-it | Anesthesiology | 2019 | 83 | 100 | 83.00 |
| medgemma-27b-text-it | Anesthesiology | 2020 | 76 | 100 | 76.00 |
| medgemma-27b-text-it | Anesthesiology | 2022 | 76 | 99 | 76.77 |
| medgemma-27b-text-it | Anesthesiology | 2023 | 77 | 100 | 77.00 |
| medgemma-27b-text-it | Anesthesiology | 2024 | 84 | 100 | 84.00 |
| medgemma-27b-text-it | Anesthesiology | 2025 | 84 | 100 | 84.00 |
| medgemma-27b-text-it | General Surgery | 2018 | 74 | 100 | 74.00 |
| medgemma-27b-text-it | General Surgery | 2019 | 77 | 100 | 77.00 |
| medgemma-27b-text-it | General Surgery | 2020 | 81 | 100 | 81.00 |
| medgemma-27b-text-it | General Surgery | 2022 | 83 | 100 | 83.00 |
| medgemma-27b-text-it | General Surgery | 2023 | 80 | 100 | 80.00 |
| medgemma-27b-text-it | General Surgery | 2024 | 74 | 100 | 74.00 |
| medgemma-27b-text-it | General Surgery | 2025 | 79 | 100 | 79.00 |
| medgemma-27b-text-it | Thoracic and Cardiovascular Surgery | 2019 | 60 | 100 | 60.00 |
| medgemma-27b-text-it | Thoracic and Cardiovascular Surgery | 2022 | 79 | 100 | 79.00 |
| medgemma-27b-text-it | Thoracic and Cardiovascular Surgery | 2023 | 82 | 100 | 82.00 |
| medgemma-27b-text-it | Thoracic and Cardiovascular Surgery | 2024 | 81 | 100 | 81.00 |
| medgemma-27b-text-it | Thoracic and Cardiovascular Surgery | 2025 | 158 | 200 | 79.00 |
| medgemma-27b-text-it | Gynecology and Obstetrics | 2018 | 77 | 100 | 77.00 |
| medgemma-27b-text-it | Gynecology and Obstetrics | 2019 | 83 | 100 | 83.00 |
| medgemma-27b-text-it | Gynecology and Obstetrics | 2020 | 82 | 100 | 82.00 |
| medgemma-27b-text-it | Gynecology and Obstetrics | 2022 | 80 | 100 | 80.00 |
| medgemma-27b-text-it | Gynecology and Obstetrics | 2023 | 79 | 100 | 79.00 |
| medgemma-27b-text-it | Gynecology and Obstetrics | 2024 | 80 | 100 | 80.00 |
| medgemma-27b-text-it | Gynecology and Obstetrics | 2025 | 83 | 100 | 83.00 |
| medgemma-27b-text-it | Neurosurgery | 2022 | 71 | 100 | 71.00 |
| medgemma-27b-text-it | Neurosurgery | 2023 | 87 | 100 | 87.00 |
| medgemma-27b-text-it | Neurosurgery | 2024 | 83 | 100 | 83.00 |
| medgemma-27b-text-it | Neurosurgery | 2025 | 68 | 100 | 68.00 |
| medgemma-27b-text-it | Ophthalmology | 2018 | 75 | 100 | 75.00 |
| medgemma-27b-text-it | Ophthalmology | 2019 | 78 | 100 | 78.00 |
| medgemma-27b-text-it | Ophthalmology | 2020 | 80 | 100 | 80.00 |
| medgemma-27b-text-it | Ophthalmology | 2022 | 75 | 100 | 75.00 |
| medgemma-27b-text-it | Ophthalmology | 2023 | 76 | 100 | 76.00 |
| medgemma-27b-text-it | Ophthalmology | 2024 | 74 | 100 | 74.00 |
| medgemma-27b-text-it | Ophthalmology | 2025 | 77 | 100 | 77.00 |
| medgemma-27b-text-it | Pediatrics | 2018 | 78 | 100 | 78.00 |
| medgemma-27b-text-it | Pediatrics | 2019 | 91 | 100 | 91.00 |
| medgemma-27b-text-it | Pediatrics | 2020 | 74 | 100 | 74.00 |
| medgemma-27b-text-it | Pediatrics | 2022 | 85 | 100 | 85.00 |
| medgemma-27b-text-it | Pediatrics | 2023 | 85 | 100 | 85.00 |
| medgemma-27b-text-it | Pediatrics | 2024 | 86 | 100 | 86.00 |
| medgemma-27b-text-it | Pediatrics | 2025 | 75 | 100 | 75.00 |
| medgemma-27b-text-it | Test A | 2018 | 83 | 100 | 83.00 |
| medgemma-27b-text-it | Test A | 2019 | 82 | 100 | 82.00 |
| medgemma-27b-text-it | Test A | 2020 | 74 | 90 | 82.22 |
| medgemma-27b-text-it | Test A | 2022 | 87 | 100 | 87.00 |
| medgemma-27b-text-it | Test A | 2023 | 84 | 100 | 84.00 |
| medgemma-27b-text-it | Test A | 2024 | 91 | 100 | 91.00 |
| medgemma-27b-text-it | Test A | 2025 | 82 | 100 | 82.00 |
| medgemma-27b-text-it | Test B | 2018 | 91 | 100 | 91.00 |
| medgemma-27b-text-it | Test B | 2019 | 86 | 100 | 86.00 |
| medgemma-27b-text-it | Test B | 2020 | 82 | 90 | 91.11 |
| medgemma-27b-text-it | Test B | 2022 | 87 | 100 | 87.00 |
| medgemma-27b-text-it | Test B | 2023 | 87 | 100 | 87.00 |
| medgemma-27b-text-it | Test B | 2024 | 86 | 100 | 86.00 |
| medgemma-27b-text-it | Test B | 2025 | 84 | 100 | 84.00 |
| medgemma-27b-text-it | Psychiatry | 2018 | 85 | 100 | 85.00 |
| medgemma-27b-text-it | Psychiatry | 2019 | 89 | 100 | 89.00 |
| medgemma-27b-text-it | Psychiatry | 2020 | 90 | 100 | 90.00 |
| medgemma-27b-text-it | Psychiatry | 2022 | 89 | 100 | 89.00 |
| medgemma-27b-text-it | Psychiatry | 2023 | 88 | 100 | 88.00 |
| medgemma-27b-text-it | Psychiatry | 2024 | 90 | 100 | 90.00 |
| medgemma-27b-text-it | Psychiatry | 2025 | 94 | 100 | 94.00 |
| medgemma-27b-text-it | Radiology | 2018 | 82 | 100 | 82.00 |
| medgemma-27b-text-it | Radiology | 2019 | 79 | 100 | 79.00 |
| medgemma-27b-text-it | Radiology | 2022 | 69 | 100 | 69.00 |
| medgemma-27b-text-it | Radiology | 2023 | 78 | 99 | 78.79 |
| medgemma-27b-text-it | Radiology | 2024 | 76 | 100 | 76.00 |
| medgemma-27b-text-it | Radiology | 2025 | 72 | 100 | 72.00 |
| medgemma-27b-text-it | Urology | 2018 | 86 | 100 | 86.00 |
| medgemma-27b-text-it | Urology | 2020 | 78 | 100 | 78.00 |
| medgemma-27b-text-it | Urology | 2022 | 82 | 100 | 82.00 |
| medgemma-27b-text-it | Urology | 2023 | 83 | 100 | 83.00 |
| medgemma-27b-text-it | Urology | 2024 | 72 | 100 | 72.00 |
| medgemma-27b-text-it | Urology | 2025 | 75 | 100 | 75.00 |
| OctoMed-7B | Pathological Anatomy and Pathology | 2018 | 79 | 100 | 79.00 |
| OctoMed-7B | Pathological Anatomy and Pathology | 2019 | 75 | 100 | 75.00 |
| OctoMed-7B | Pathological Anatomy and Pathology | 2020 | 83 | 100 | 83.00 |
| OctoMed-7B | Pathological Anatomy and Pathology | 2023 | 82 | 100 | 82.00 |
| OctoMed-7B | Pathological Anatomy and Pathology | 2024 | 81 | 100 | 81.00 |
| OctoMed-7B | Pathological Anatomy and Pathology | 2025 | 74 | 100 | 74.00 |
| OctoMed-7B | Anesthesiology | 2018 | 79 | 100 | 79.00 |
| OctoMed-7B | Anesthesiology | 2019 | 79 | 100 | 79.00 |
| OctoMed-7B | Anesthesiology | 2020 | 74 | 98 | 75.51 |
| OctoMed-7B | Anesthesiology | 2022 | 81 | 99 | 81.82 |
| OctoMed-7B | Anesthesiology | 2023 | 65 | 99 | 65.66 |
| OctoMed-7B | Anesthesiology | 2024 | 75 | 100 | 75.00 |
| OctoMed-7B | Anesthesiology | 2025 | 80 | 100 | 80.00 |
| OctoMed-7B | General Surgery | 2018 | 67 | 99 | 67.68 |
| OctoMed-7B | General Surgery | 2019 | 79 | 100 | 79.00 |
| OctoMed-7B | General Surgery | 2020 | 72 | 100 | 72.00 |
| OctoMed-7B | General Surgery | 2022 | 77 | 100 | 77.00 |
| OctoMed-7B | General Surgery | 2023 | 83 | 100 | 83.00 |
| OctoMed-7B | General Surgery | 2024 | 67 | 100 | 67.00 |
| OctoMed-7B | General Surgery | 2025 | 73 | 100 | 73.00 |
| OctoMed-7B | Thoracic and Cardiovascular Surgery | 2019 | 60 | 100 | 60.00 |
| OctoMed-7B | Thoracic and Cardiovascular Surgery | 2022 | 74 | 100 | 74.00 |
| OctoMed-7B | Thoracic and Cardiovascular Surgery | 2023 | 79 | 100 | 79.00 |
| OctoMed-7B | Thoracic and Cardiovascular Surgery | 2024 | 72 | 100 | 72.00 |
| OctoMed-7B | Thoracic and Cardiovascular Surgery | 2025 | 142 | 200 | 71.00 |
| OctoMed-7B | Gynecology and Obstetrics | 2018 | 85 | 99 | 85.86 |
| OctoMed-7B | Gynecology and Obstetrics | 2019 | 83 | 100 | 83.00 |
| OctoMed-7B | Gynecology and Obstetrics | 2020 | 82 | 100 | 82.00 |
| OctoMed-7B | Gynecology and Obstetrics | 2022 | 73 | 100 | 73.00 |
| OctoMed-7B | Gynecology and Obstetrics | 2023 | 80 | 100 | 80.00 |
| OctoMed-7B | Gynecology and Obstetrics | 2024 | 71 | 100 | 71.00 |
| OctoMed-7B | Gynecology and Obstetrics | 2025 | 72 | 100 | 72.00 |
| OctoMed-7B | Neurosurgery | 2022 | 74 | 100 | 74.00 |
| OctoMed-7B | Neurosurgery | 2023 | 85 | 100 | 85.00 |
| OctoMed-7B | Neurosurgery | 2024 | 78 | 99 | 78.79 |
| OctoMed-7B | Neurosurgery | 2025 | 71 | 99 | 71.72 |
| OctoMed-7B | Ophthalmology | 2018 | 76 | 100 | 76.00 |
| OctoMed-7B | Ophthalmology | 2019 | 68 | 100 | 68.00 |
| OctoMed-7B | Ophthalmology | 2020 | 78 | 100 | 78.00 |
| OctoMed-7B | Ophthalmology | 2022 | 74 | 100 | 74.00 |
| OctoMed-7B | Ophthalmology | 2023 | 72 | 100 | 72.00 |
| OctoMed-7B | Ophthalmology | 2024 | 68 | 99 | 68.69 |
| OctoMed-7B | Ophthalmology | 2025 | 76 | 100 | 76.00 |
| OctoMed-7B | Pediatrics | 2018 | 81 | 100 | 81.00 |
| OctoMed-7B | Pediatrics | 2019 | 82 | 100 | 82.00 |
| OctoMed-7B | Pediatrics | 2020 | 70 | 100 | 70.00 |
| OctoMed-7B | Pediatrics | 2022 | 77 | 100 | 77.00 |
| OctoMed-7B | Pediatrics | 2023 | 71 | 99 | 71.72 |
| OctoMed-7B | Pediatrics | 2024 | 78 | 100 | 78.00 |
| OctoMed-7B | Pediatrics | 2025 | 79 | 100 | 79.00 |
| OctoMed-7B | Test A | 2018 | 84 | 100 | 84.00 |
| OctoMed-7B | Test A | 2019 | 80 | 100 | 80.00 |
| OctoMed-7B | Test A | 2020 | 77 | 90 | 85.56 |
| OctoMed-7B | Test A | 2022 | 76 | 100 | 76.00 |
| OctoMed-7B | Test A | 2023 | 84 | 100 | 84.00 |
| OctoMed-7B | Test A | 2024 | 84 | 100 | 84.00 |
| OctoMed-7B | Test A | 2025 | 80 | 100 | 80.00 |
| OctoMed-7B | Test B | 2018 | 86 | 100 | 86.00 |
| OctoMed-7B | Test B | 2019 | 83 | 100 | 83.00 |
| OctoMed-7B | Test B | 2020 | 79 | 90 | 87.78 |
| OctoMed-7B | Test B | 2022 | 81 | 99 | 81.82 |
| OctoMed-7B | Test B | 2023 | 89 | 99 | 89.90 |
| OctoMed-7B | Test B | 2024 | 82 | 100 | 82.00 |
| OctoMed-7B | Test B | 2025 | 83 | 99 | 83.84 |
| OctoMed-7B | Psychiatry | 2018 | 79 | 100 | 79.00 |
| OctoMed-7B | Psychiatry | 2019 | 88 | 100 | 88.00 |
| OctoMed-7B | Psychiatry | 2020 | 87 | 100 | 87.00 |
| OctoMed-7B | Psychiatry | 2022 | 87 | 100 | 87.00 |
| OctoMed-7B | Psychiatry | 2023 | 89 | 100 | 89.00 |
| OctoMed-7B | Psychiatry | 2024 | 78 | 100 | 78.00 |
| OctoMed-7B | Psychiatry | 2025 | 89 | 100 | 89.00 |
| OctoMed-7B | Radiology | 2018 | 76 | 99 | 76.77 |
| OctoMed-7B | Radiology | 2019 | 84 | 100 | 84.00 |
| OctoMed-7B | Radiology | 2022 | 70 | 100 | 70.00 |
| OctoMed-7B | Radiology | 2023 | 84 | 100 | 84.00 |
| OctoMed-7B | Radiology | 2024 | 72 | 100 | 72.00 |
| OctoMed-7B | Radiology | 2025 | 77 | 100 | 77.00 |
| OctoMed-7B | Urology | 2018 | 79 | 99 | 79.80 |
| OctoMed-7B | Urology | 2020 | 73 | 100 | 73.00 |
| OctoMed-7B | Urology | 2022 | 81 | 100 | 81.00 |
| OctoMed-7B | Urology | 2023 | 79 | 100 | 79.00 |
| OctoMed-7B | Urology | 2024 | 71 | 100 | 71.00 |
| OctoMed-7B | Urology | 2025 | 79 | 100 | 79.00 |
| medgemma-4b-it-FT | Pathological Anatomy and Pathology | 2018 | 67 | 100 | 67.00 |
| medgemma-4b-it-FT | Pathological Anatomy and Pathology | 2019 | 67 | 100 | 67.00 |
| medgemma-4b-it-FT | Pathological Anatomy and Pathology | 2020 | 66 | 100 | 66.00 |
| medgemma-4b-it-FT | Pathological Anatomy and Pathology | 2023 | 68 | 100 | 68.00 |
| medgemma-4b-it-FT | Pathological Anatomy and Pathology | 2024 | 75 | 100 | 75.00 |
| medgemma-4b-it-FT | Pathological Anatomy and Pathology | 2025 | 50 | 100 | 50.00 |
| medgemma-4b-it-FT | Anesthesiology | 2018 | 63 | 100 | 63.00 |
| medgemma-4b-it-FT | Anesthesiology | 2019 | 69 | 100 | 69.00 |
| medgemma-4b-it-FT | Anesthesiology | 2020 | 64 | 100 | 64.00 |
| medgemma-4b-it-FT | Anesthesiology | 2022 | 67 | 100 | 67.00 |
| medgemma-4b-it-FT | Anesthesiology | 2023 | 63 | 100 | 63.00 |
| medgemma-4b-it-FT | Anesthesiology | 2024 | 78 | 100 | 78.00 |
| medgemma-4b-it-FT | Anesthesiology | 2025 | 58 | 100 | 58.00 |
| medgemma-4b-it-FT | General Surgery | 2018 | 62 | 100 | 62.00 |
| medgemma-4b-it-FT | General Surgery | 2019 | 65 | 100 | 65.00 |
| medgemma-4b-it-FT | General Surgery | 2020 | 71 | 100 | 71.00 |
| medgemma-4b-it-FT | General Surgery | 2022 | 67 | 100 | 67.00 |
| medgemma-4b-it-FT | General Surgery | 2023 | 76 | 100 | 76.00 |
| medgemma-4b-it-FT | General Surgery | 2024 | 58 | 100 | 58.00 |
| medgemma-4b-it-FT | General Surgery | 2025 | 62 | 100 | 62.00 |
| medgemma-4b-it-FT | Thoracic and Cardiovascular Surgery | 2019 | 59 | 100 | 59.00 |
| medgemma-4b-it-FT | Thoracic and Cardiovascular Surgery | 2022 | 77 | 100 | 77.00 |
| medgemma-4b-it-FT | Thoracic and Cardiovascular Surgery | 2023 | 78 | 100 | 78.00 |
| medgemma-4b-it-FT | Thoracic and Cardiovascular Surgery | 2024 | 75 | 100 | 75.00 |
| medgemma-4b-it-FT | Thoracic and Cardiovascular Surgery | 2025 | 102 | 200 | 51.00 |
| medgemma-4b-it-FT | Gynecology and Obstetrics | 2018 | 64 | 100 | 64.00 |
| medgemma-4b-it-FT | Gynecology and Obstetrics | 2019 | 71 | 100 | 71.00 |
| medgemma-4b-it-FT | Gynecology and Obstetrics | 2020 | 61 | 100 | 61.00 |
| medgemma-4b-it-FT | Gynecology and Obstetrics | 2022 | 66 | 100 | 66.00 |
| medgemma-4b-it-FT | Gynecology and Obstetrics | 2023 | 61 | 100 | 61.00 |
| medgemma-4b-it-FT | Gynecology and Obstetrics | 2024 | 68 | 100 | 68.00 |
| medgemma-4b-it-FT | Gynecology and Obstetrics | 2025 | 51 | 100 | 51.00 |
| medgemma-4b-it-FT | Neurosurgery | 2022 | 58 | 100 | 58.00 |
| medgemma-4b-it-FT | Neurosurgery | 2023 | 70 | 100 | 70.00 |
| medgemma-4b-it-FT | Neurosurgery | 2024 | 75 | 100 | 75.00 |
| medgemma-4b-it-FT | Neurosurgery | 2025 | 45 | 100 | 45.00 |
| medgemma-4b-it-FT | Ophthalmology | 2018 | 50 | 100 | 50.00 |
| medgemma-4b-it-FT | Ophthalmology | 2019 | 56 | 100 | 56.00 |
| medgemma-4b-it-FT | Ophthalmology | 2020 | 71 | 100 | 71.00 |
| medgemma-4b-it-FT | Ophthalmology | 2022 | 59 | 100 | 59.00 |
| medgemma-4b-it-FT | Ophthalmology | 2023 | 60 | 100 | 60.00 |
| medgemma-4b-it-FT | Ophthalmology | 2024 | 53 | 100 | 53.00 |
| medgemma-4b-it-FT | Ophthalmology | 2025 | 42 | 100 | 42.00 |
| medgemma-4b-it-FT | Pediatrics | 2018 | 73 | 100 | 73.00 |
| medgemma-4b-it-FT | Pediatrics | 2019 | 79 | 100 | 79.00 |
| medgemma-4b-it-FT | Pediatrics | 2020 | 73 | 100 | 73.00 |
| medgemma-4b-it-FT | Pediatrics | 2022 | 86 | 100 | 86.00 |
| medgemma-4b-it-FT | Pediatrics | 2023 | 77 | 100 | 77.00 |
| medgemma-4b-it-FT | Pediatrics | 2024 | 83 | 100 | 83.00 |
| medgemma-4b-it-FT | Pediatrics | 2025 | 50 | 100 | 50.00 |
| medgemma-4b-it-FT | Test A | 2018 | 73 | 100 | 73.00 |
| medgemma-4b-it-FT | Test A | 2019 | 83 | 100 | 83.00 |
| medgemma-4b-it-FT | Test A | 2020 | 62 | 90 | 68.89 |
| medgemma-4b-it-FT | Test A | 2022 | 74 | 100 | 74.00 |
| medgemma-4b-it-FT | Test A | 2023 | 86 | 100 | 86.00 |
| medgemma-4b-it-FT | Test A | 2024 | 80 | 100 | 80.00 |
| medgemma-4b-it-FT | Test A | 2025 | 64 | 100 | 64.00 |
| medgemma-4b-it-FT | Test B | 2018 | 71 | 100 | 71.00 |
| medgemma-4b-it-FT | Test B | 2019 | 74 | 100 | 74.00 |
| medgemma-4b-it-FT | Test B | 2020 | 70 | 90 | 77.78 |
| medgemma-4b-it-FT | Test B | 2022 | 73 | 100 | 73.00 |
| medgemma-4b-it-FT | Test B | 2023 | 65 | 100 | 65.00 |
| medgemma-4b-it-FT | Test B | 2024 | 70 | 100 | 70.00 |
| medgemma-4b-it-FT | Test B | 2025 | 55 | 100 | 55.00 |
| medgemma-4b-it-FT | Psychiatry | 2018 | 74 | 100 | 74.00 |
| medgemma-4b-it-FT | Psychiatry | 2019 | 88 | 100 | 88.00 |
| medgemma-4b-it-FT | Psychiatry | 2020 | 87 | 100 | 87.00 |
| medgemma-4b-it-FT | Psychiatry | 2022 | 80 | 100 | 80.00 |
| medgemma-4b-it-FT | Psychiatry | 2023 | 86 | 100 | 86.00 |
| medgemma-4b-it-FT | Psychiatry | 2024 | 86 | 100 | 86.00 |
| medgemma-4b-it-FT | Psychiatry | 2025 | 78 | 100 | 78.00 |
| medgemma-4b-it-FT | Radiology | 2018 | 71 | 100 | 71.00 |
| medgemma-4b-it-FT | Radiology | 2019 | 72 | 100 | 72.00 |
| medgemma-4b-it-FT | Radiology | 2022 | 68 | 100 | 68.00 |
| medgemma-4b-it-FT | Radiology | 2023 | 72 | 100 | 72.00 |
| medgemma-4b-it-FT | Radiology | 2024 | 74 | 100 | 74.00 |
| medgemma-4b-it-FT | Radiology | 2025 | 49 | 100 | 49.00 |
| medgemma-4b-it-FT | Urology | 2018 | 71 | 100 | 71.00 |
| medgemma-4b-it-FT | Urology | 2020 | 67 | 100 | 67.00 |
| medgemma-4b-it-FT | Urology | 2022 | 68 | 100 | 68.00 |

| medgemma-4b-it-FT | Urology | 2023 | 76 | 100 | 76.00 |
|---|---|---|---|---|---|
| medgemma-4b-it-FT | Urology | 2024 | 68 | 100 | 68.00 |
| medgemma-4b-it-FT | Urology | 2025 | 44 | 100 | 44.00 |

**Supplementary Table 6. Percentage of questions that were correctly answered by each large langue (LLM) model only.**

| Model (LLM) | Percentage (absolute number) of correct answers provided by each LLM alone |
|---|---|
| medgemma-4b-it | 0.20 (17) |
| BioMistral-7B-DARE | 0.26 (22) |
| MediPhi-Instruct-3.8B | 0.44 (37) |
| Llama3-OpenBioLLM-8B | 0.18 (15) |
| JSL-MedLlama-3-8B-v2.0 | 0.17 (14) |
| meditron-7b | 0.11 (09) |
| OctoMed-7B | 0.60 (50) |
| medgemma-1.5-4b-it | 0.28 (23) |
| medgemma-27b-text-it | 0.55 (46) |
| Llama3-OpenBioLLM-70B | 0.42 (35) |
| medgemma-4b-it-FT (fine-tuned version) | 0.72 (60) |

The denominator represented the total number of questions posed (8,380). For instance, in a specific example, only medgemma-4b-it answered 0.34% of the questions correctly, while all other LLMs provided incorrect responses for those questions.

**Supplementary Table 7. Percentage of questions that none of the large language models (LLMs) answered correctly by year.**

| Year | Percentage (%) |
|---|---|
| 2025 | 21.05 |
| 2024 | 17.90 |
| 2023 | 14.21 |
| 2022 | 13.68 |
| 2020 | 7.37 |
| 2019 | 11.05 |
| 2018 | 14.74 |

The denominator was 190 questions for which none of the LLMs provided the correct answer.

**Supplementary Table 8. Percentage of questions that none of the large language models (LLMs) answered correctly by medical specialty and subspecialty.**

| Specialty and Subspecialty | Percentage (%) |
|---|---|
| Pathology and Anatomical Pathology | 7.90 |
| Anesthesiology | 6.84 |
| General Surgery | 11.05 |
| Thoracic and Cardiovascular Surgery | 14.21 |
| Gynecology & Obstetrics | 5.79 |
| Neurosurgery | 3.68 |
| Ophthalmology | 12.63 |
| Pediatrics | 7.90 |
| Psychiatry | 3.16 |
| Radiology | 8.95 |
| Urology | 8.42 |
| Test A | 4.74 |
| Test B | 4.74 |

The denominator was 190 questions for which none of the LLMs provided the correct answer.